\documentclass{article}

\usepackage[svgnames]{xcolor}
\usepackage{todonotes}

\usepackage{nac_preprint}
\usepackage[utf8]{inputenc} 
\usepackage[T1]{fontenc}    
\usepackage{hyperref}       
\usepackage{url}            
\usepackage{booktabs}       
\usepackage{amsfonts}       
\usepackage{nicefrac}       
\usepackage{microtype}      

\usepackage{booktabs} 
\usepackage{subcaption}
\usepackage{amsmath}
\usepackage{amssymb}
\usepackage{cleveref}
\usepackage{mathtools}
\usepackage{enumitem}
\usepackage{graphics}
\usepackage{graphicx}

\usepackage[export]{adjustbox}
\usepackage{algorithm}
\usepackage{algorithmicx}
\usepackage[noend]{algpseudocode}
\algnewcommand{\LineComment}[1]{\State \(//\) #1}
\algnewcommand{\RLineComment}[1]{\State \(\triangleright\) #1}
\usepackage{bm}
\usepackage{multirow}

\usepackage{wrapfig}

\DeclareMathOperator*{\argmax}{argmax}

\usepackage{etoolbox}
\usepackage{tikz}
\usetikzlibrary{tikzmark}
\usetikzlibrary{calc}

\errorcontextlines\maxdimen

\newcommand{\ALGtikzmarkcolor}{black}
\newcommand{\ALGtikzmarkextraindent}{4pt}
\newcommand{\ALGtikzmarkverticaloffsetstart}{-.5ex}
\newcommand{\ALGtikzmarkverticaloffsetend}{-.5ex}
\makeatletter
\newcounter{ALG@tikzmark@tempcnta}

\newcommand\ALG@tikzmark@start{%
    \global\let\ALG@tikzmark@last\ALG@tikzmark@starttext%
    \expandafter\edef\csname ALG@tikzmark@\theALG@nested\endcsname{\theALG@tikzmark@tempcnta}%
    \tikzmark{ALG@tikzmark@start@\csname ALG@tikzmark@\theALG@nested\endcsname}%
    \addtocounter{ALG@tikzmark@tempcnta}{1}%
}

\def\ALG@tikzmark@starttext{start}
\newcommand\ALG@tikzmark@end{%
    \ifx\ALG@tikzmark@last\ALG@tikzmark@starttext
    \else
        \tikzmark{ALG@tikzmark@end@\csname ALG@tikzmark@\theALG@nested\endcsname}%
        \tikz[overlay,remember picture] \draw[\ALGtikzmarkcolor] let \p{S}=($(pic cs:ALG@tikzmark@start@\csname ALG@tikzmark@\theALG@nested\endcsname)+(\ALGtikzmarkextraindent,\ALGtikzmarkverticaloffsetstart)$), \p{E}=($(pic cs:ALG@tikzmark@end@\csname ALG@tikzmark@\theALG@nested\endcsname)+(\ALGtikzmarkextraindent,\ALGtikzmarkverticaloffsetend)$) in (\x{S},\y{S})--(\x{S},\y{E});%
    \fi
    \gdef\ALG@tikzmark@last{end}%
}

\apptocmd{\ALG@beginblock}{\ALG@tikzmark@start}{}{\errmessage{failed to patch}}
\pretocmd{\ALG@endblock}{\ALG@tikzmark@end}{}{\errmessage{failed to patch}}
\makeatother

\usepackage{pifont}
\newcommand{\cmark}{\textcolor{teal}{\ding{52}}}%
\newcommand{\xmark}{\textcolor{red}{\ding{55}}}%

\title{Contrastive-Signal-Dependent Plasticity: Self-Supervised Learning {in} Spiking Neural Circuits}

\author{%
  Alexander Ororbia \\
  Rochester Institute of Technology \\
  \texttt{ago@cs.rit.edu}
}

\begin{document}

\setlength{\abovedisplayskip}{0.065cm}
\setlength{\belowdisplayskip}{0pt}

\maketitle

\begin{abstract}
Brain-inspired machine intelligence research seeks to develop computational models that emulate the information processing and adaptability that {distinguishes} biological systems of neurons. This has led to the development of spiking neural networks, a class of models that promisingly addresses the biological implausibility and {the lack of energy efficiency} inherent to modern-day deep neural networks. In this work, we address the challenge of designing neurobiologically-motivated schemes for adjusting the synapses of spiking networks and propose \emph{contrastive-signal-dependent plasticity}, a process which generalizes ideas behind self-supervised learning to facilitate local adaptation in architectures of event-based neuronal layers that operate in parallel. Our experimental simulations demonstrate a consistent advantage over other biologically-plausible approaches when training recurrent spiking networks, crucially side-stepping the need for extra structure such as feedback synapses.\footnote{This paper was published in Science Advances. Please cite the following:  \\
Alexander G. Ororbia,
Contrastive signal–dependent plasticity: Self-supervised learning in spiking neural circuits. Sci. Adv.10, eadn6076 (2024). DOI: 10.1126/sciadv.adn6076} 

\small{
}
\end{abstract}

\section{Introduction}
\label{sec:intro}

The notion of `mortal computation' \cite{hinton2022forward,ororbia2022predictive,ororbia2023mortal} challenges one of the foundational principles upon which general purpose computers have been built. Specifically, computation, as it is conducted today, is driven by the strong separation of software from hardware{. This} means that the knowledge contained within a program written in the software is ``immortal'', allowing it to be copied to different physical copies of the hardware itself. Machine learning models, which can be viewed as programs that adjust themselves in accordance with data, also rely upon this separation. Mortal computation, in contrast, means that once the hardware medium fails or ``dies'', the information encoded within it will also disappear, much akin to what would happen to the knowledge acquired by a biological organism when it is no longer able to maintain homeostasis. Despite the transience that comes with the 
{binding} of software to hardware, a valuable property emerges -- energy efficiency. 
This motivates a move away from \emph{Red AI}, the result of immortal computation, {and towards} \emph{Green AI} \cite{schwartz2020green}, addressing a key concern: how {might} we design intelligent systems that do not escalate computational and carbon costs \cite{brevini2020black,schwartz2020green,patterson2021carbon}? 

A promising pathway to mortal computation centers around a family of brain-inspired computational models known as spiking neural networks (SNNs) \cite{maass1997networks,eliasmith2012large,gutig2016spiking,luo2021architectures} -- neuronal `software' that processes and transmits information via discrete electrical pulses -- in tandem with hardware such as memristor systems \cite{itoh2008memristor,thomas2013memristor} and neuromorphic chips \cite{merolla2014million,roy2019towards}. In particular, it has been shown that neuromorphic-instantiated SNNs can be several orders of magnitude more power efficient \cite{wang2018learning,pei2019towards} than modern-day deep neural networks \cite{hinton2006reducing}{. This} type of in silico mortal computation could bring to machine intelligence elements of the human brain's {comparatively} minimal energy consumption \cite{cox2014neural}. 
However, a challenge facing the development of these types of systems is the design of the requisite architectures and `credit assignment' schemes, i.e., algorithms that determine the positive/negative contributions of each neuron to an SNN's overall behavioral improvement, that could be cast in terms of in-memory processing \cite{zou2021breaking,rao2021homogeneous} 
-- where memories correspond to the SNN's synaptic connections -- in order to side-step the ‘von Neumann bottleneck’ \cite{zou2021breaking}, or the extra thermodynamic costs induced by reading and writing to memory. 
In service of the efficiency afforded by a neuromorphic mortal computer, some of the key qualities that should characterize its underlying inference and learning computations include:
\textbf{1)} no requirement for differentiability in the spike-based communication (backprop through time as applied to SNNs requires differentiability and thus the careful design of surrogate functions \cite{lee2016training,yin2017algorithm}),  
\textbf{2)} no forward-locking \cite{jaderberg2017decoupled} in the propagation of information through the system (a layer of neurons can compute their activities without waiting on other layers to update their own), 
\textbf{3)} no backward-locking \cite{jaderberg2017decoupled} in the synaptic updating phase (changes to synapses for one layer of neurons can be executed in parallel with other layers across time -- updates are local in both space and time), and 
\textbf{4)} {no requirement for feedback synaptic pathways} \cite{salvatori2023brain}, especially those that traverse backwards along the same forward propagation pathways (weight transport) \cite{grossberg1987competitive}, to calculate synaptic change (frameworks such as spiking predictive coding require feedback pathways to carry out the requisite message passing \cite{rao2004hierarchical,ororbia2019spiking,ndri2023}). 

In this study, we satisfy the above {criteria} through the development of a biologically-plausible neural circuit and a self-supervised learning scheme that formulates forward-only \cite{kohan2018error} and forward-forward (FF) credit assignment \cite{hinton2022forward,ororbia2022predictive} for spike-based neuronal communication. {Forward-forward (FF) learning, in general, formulates the adaptation of the  parameters of an artificial neural network (ANN) as a process of manipulating the probabilities that it assigns to the data points presented to it, i.e., the goal is for the ANN to raise the probability for actual data while lowering the probability for fake data. Adjustments are then made, under FF, to the synaptic weights of the ANN based on these probabilities in the context of a scoring objective assigned to each of its layers. The principles underlying FF adaptation motivate} 
this work's central contributions, {which} include: 
\begin{enumerate}[noitemsep,nolistsep]
    \item the design of a recurrent spiking neural circuit that exhibits layer-wise parallelism, where each layer is driven by top-down, bottom-up, and lateral pressures that do not require feedback synaptic pathways across the network -- this means that our learning and inference processes directly resolve the forward and update locking problems, 
    \item the proposal of \emph{contrastive-signal-dependent plasticity} (CSDP) for adapting the synapses of spiking neural systems in a dynamic fashion suitable for in-memory computation, further offering a complementary rule to spike-timing dependent plasticity (STDP) \cite{bi1998synaptic}, 
    \item the development of simple, fast neural circuit mechanisms that locally learn to classify and/or reconstruct activity signals, 
    and 
    \item {a} quantitative evaluation of the generalization ability of our CSDP-learned {SNNs}.
\end{enumerate}

\section*{Results}

\begin{figure}[!t]
  \begin{center}
    \includegraphics[width=0.35\textwidth]{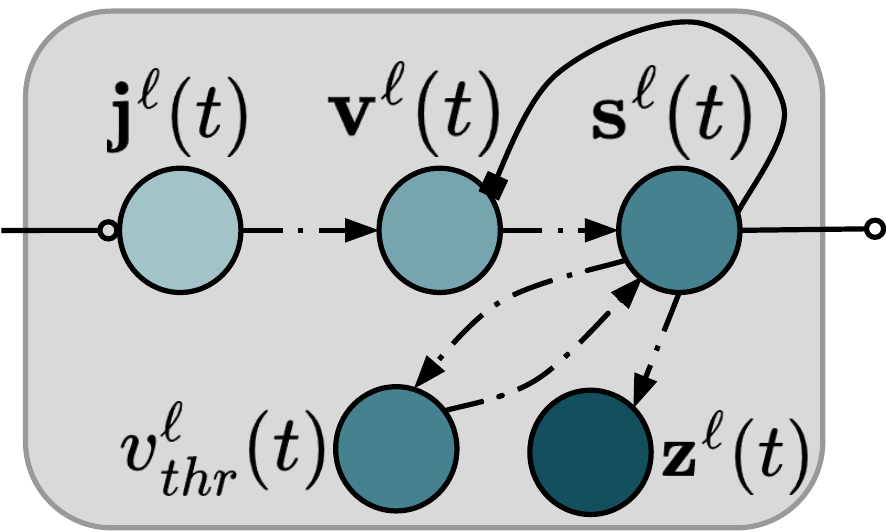}
  \end{center}
  \vspace{-0.2cm}
  \caption{ \textbf{{Componential} layout of one layer in the {CSDP SNN} recurrent circuit.} 
  A {component} diagram of a {leaky integrate-and-fire (LIF)} cell group/vector. The neuronal {components} include the electrical current $\mathbf{j}^\ell(t)$, the membrane voltage potential $\mathbf{v}^\ell(t)$, the spike emission $\mathbf{s}^\ell(t)$, the activation trace $\mathbf{z}^\ell(t)$, and the cross-layer homeostatically-constrained threshold $v^\ell_{thr}(t)$. {Black (dashed) arrows with solid triangle heads indicate the flow of values within an LIF group's dynamics (e.g.., current $\mathbf{j}^\ell$ is input to dynamics of $\mathbf{v}^\ell(t)$), open-circle heads indicate excitation/additive pressure, and solid square heads indicate inhibitory/subtractive pressure.
  }
  }
  \label{fig:lif_cell}
\end{figure}

\subsection*{A Recurrent Spiking Neural Circuit} 
The neuro-mimetic architecture designed in this work is made up of recurrent layers that operate in parallel with each other ($L$ layers in total){, where each layer $\ell$ contains $J_\ell$ neuronal cells, which {are} further made up of multiple {components} (see Fig. 1
)}. A natural by-product of our neural circuit design is that all layers of spiking neuronal cells can be simulated, at each step in time, completely in parallel, further permitting possible implementations with neural layers that are executed asynchronously, e.g., such as in coupled neuromorphic crossbars. Specifically, at time $t$, each layer $\ell$ {of neurons} takes in as input the previous (spike) signals emitted (at time $t-1$) from the neurons immediately below (from layer $\ell-1$){, i.e., } $\mathbf{s}^{\ell-1}(t) \in \{0,1\}^{J_{\ell-1} \times 1}$ {,} as well as those immediately above (from layer $\ell+1$){, i.e., } $\mathbf{s}^{\ell+1}(t) \in \{0,1\}^{J_{\ell+1} \times 1}$. $\mathbf{W}^\ell$ is the bundle of synaptic connections that propagate information from the layer below ($\ell-1$) and $\mathbf{V}^\ell$ is the synaptic bundle that transmits information from the layer above ($\ell+1$). Furthermore, the neurons in each layer are laterally connected to one another, where the synapses $\mathbf{M}^\ell$ that connect them enforce a form of dynamic inhibition{:} neurons that are more strongly active will suppress the activities of the more weakly active ones in $\mathbf{s}^\ell(t)${. This} results in an emergent form of $K$ winners-take-all competition, where $K$ changes with time. Finally, we explore the integration of a set of optional top-down ``attentional'' {class}-mediating (spike) signals $\mathbf{s}_y \in \{0,1\}^{C \times 1}$ {(where $C$ is the number of classes)}, transmitted along the synaptic bundle $\mathbf{B}^\ell${. These} encourage spiking neural units to form representations that encode any available discriminative information, e.g., labels/annotations. Incorporating these {class} contextual synapses results in a supervised model. 

Concretely, the dynamics of any single layer of neuronal processing units (NPUs) or cells within our biomimetic model follow that of the leaky integrator \cite{lapicque1907recherches} and the full dynamics (see Fig. 2 
for a visual depiction)
of any layer of spiking NPUs is calculated in terms of electrical current $\mathbf{j}^\ell(t)$ as follows:
\begin{equation}
 d^\ell_i(t) = R_E \big( \Sigma^{J_y}_{j=1} W^\ell_{ij} s^{\ell-1}_j(t)\big) + R_E \big( \Sigma^{J_{\ell+1}}_{j=1} 
 V^\ell_{ij} s^{\ell+1}_j(t) \big) - R_I \big( \Sigma^{J_\ell}_{j=1} (M^\ell_{ij} (1 - I^\ell_{ij})) \cdot s^\ell_j(t) \big)
 \end{equation}
\begin{equation}
    j^\ell_i(t) = 
    \begin{dcases}
        d^\ell_i(t) & \mathbf{y} = \emptyset \\
        d^\ell_i(t) + R_E \big( \Sigma^{J_y}_{j=1} B^\ell_{ij} \cdot s_{y,j}(t) \big) & \mathbf{y} \neq \emptyset \\ 
    \end{dcases} 
    \label{eqn:current}
\end{equation}
which then triggers an update to the NPUs' membrane voltage values $\mathbf{v}^\ell(t)$:
\begin{equation}
    \widehat{v}^\ell_i(t + \Delta t) = v^\ell_i(t) + (\Delta t/\tau_m) \big(-v^\ell_i(t) + j^\ell_i(t) \big) \label{eqn:voltage}
\end{equation}
and, finally, results in the emission of spikes $\mathbf{s}^\ell(t)$ (or action potentials):
\begin{equation} 
    s^\ell_i(t+\Delta t) = \widehat{v}^\ell_i(t + \Delta t) > v^\ell_{thr}, \; \mbox{and, } \; v^\ell_i(t + \Delta t) = \widehat{v}^\ell_i(t + \Delta t) (1 - s^\ell_i(t + \Delta t) ) \label{eqn:spike_model} 
\end{equation}
\begin{equation}
    v^\ell_{thr} = v^\ell_{thr} + \lambda_v \Big( (\Sigma^{J_\ell}_{j=1} s^\ell_j(t+\Delta t)) - 1 \Big)   \label{eqn:threshold}
\end{equation}
where 
$(1 - \mathbf{I}^\ell)$ creates a $J_\ell \times J_\ell$ hollow matrix. 
{The hollow matrix is used to ensure that there is only cross-inhibition in any layer $\ell$ and no self-excitation/inhibition (if one wanted to explicitly model self-inhibitory/excitatory effects, one could either remove the $(1 - \mathbf{I}^\ell$) to place self-inhibition or add an extra term $+ \alpha \mathbf{I}$ to incorporate self-excitation; we leave this for future work).} 
Equation 2 
states that, for any layer $\ell$ at time $t$, the electrical current input $\mathbf{j}^\ell(t)$ is a function of spatially near (i.e., bottom-up, top-down, lateral) as well as (possibly existent) {target class}-mediating action potentials. In Equation 3, 
$\mathbf{v}^\ell(t)$, the vector containing the membrane voltage potential values for each cell, is driven by the current produced by Equation 2, 
(where excitatory currents are weighted by resistance constant $R_E$ in (deci)Ohms and inhibitory currents are weighted by inhibitory resistance $R_I$){. This} voltage ultimately triggers the emission of a layer's output spike $\mathbf{s}^\ell(t)$, as in Equation 4. 
$\Delta t$ is the integration time constant on the order of milliseconds (ms) while $\tau_m$ is the membrane time constant (in ms). Further notice that our spike emission model (Equation 4
) entails a depolarization of the membrane potential{:} this (re)sets the voltage to a resting potential of $0$ (deci)volts through binary gating{. Note}
that, while we omit modeling {refractoriness} 
for simplicity, our dynamics are flexible enough to warrant incorporating both relative and absolute refractory periods to promote additional sparsity. 

\begin{figure}[!t]
     \centering
     \includegraphics[width=0.85\textwidth]{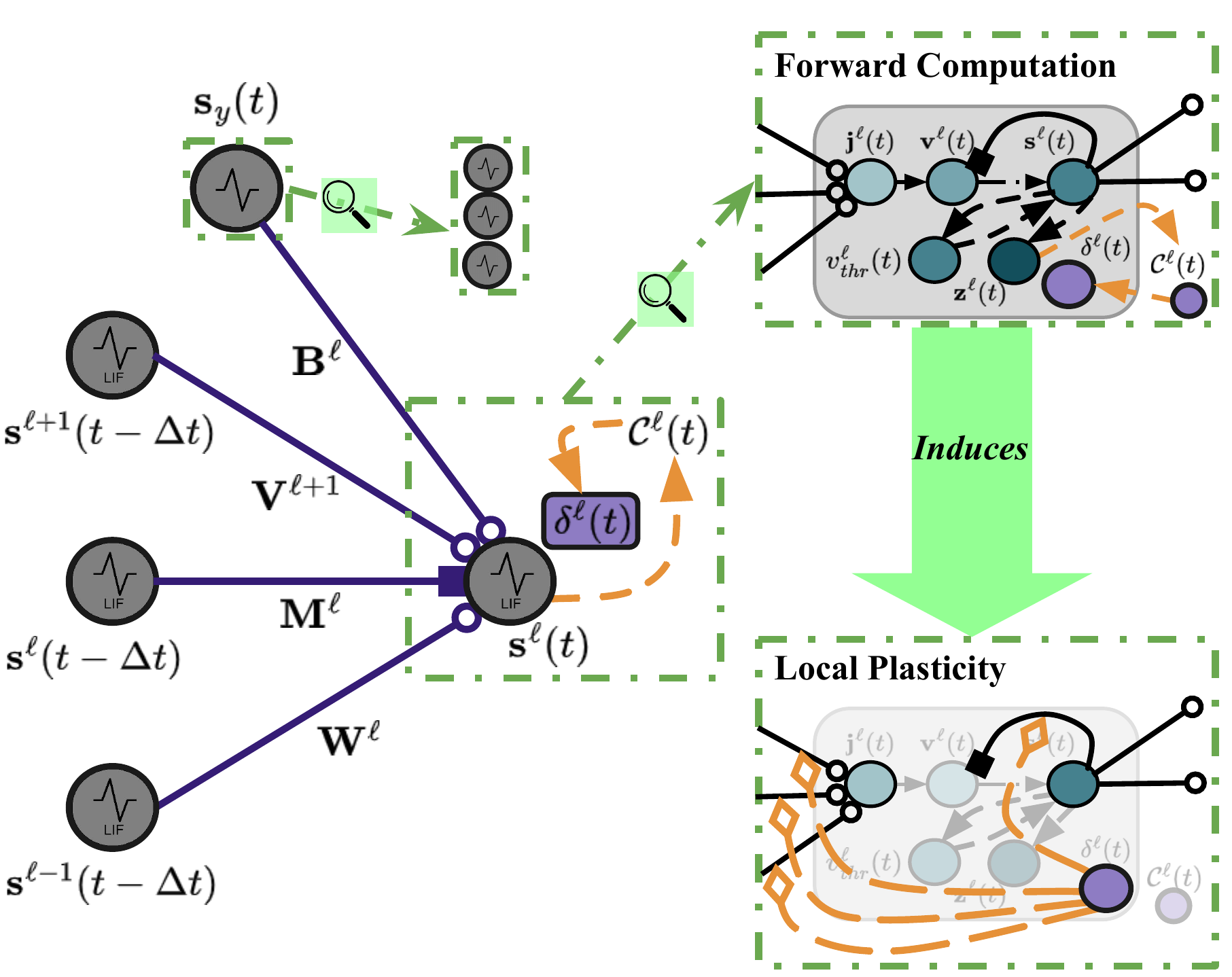}
     \vspace{-0.1cm}
        \caption{ \textbf{Neuronal layer within a CSDP-trained recurrent SNN.} 
        Depicted is the neural computation underlying one layer of our recurrent spiking circuit as well as its synaptic adjustment, induced via contrastive-signal-dependent plasticity. {Note that the supervised variant of CSDP uses class-modulation synapses $\mathbf{B}^\ell$ whereas the unsupervised variant does not.} 
        To facilitate contrastive learning, we integrate another biochemical {component} $\mathbf{\delta}^\ell(t)$ into our leaky integrator cell model. Once electrical current $\mathbf{j}^\ell(t)$ has been injected into the cells, triggering an update to their membrane potentials $\mathbf{v}^\ell(t)$, possibly leading to emission of action potentials $\mathbf{s}^\ell(t)$, their glutamate traces contribute to a goodness modulator $\mathcal{C}^\ell(t)$ which then deposits a local synaptic adjustment signal to each cell's $\delta^\ell_i(t)$ {component}. Solid lines indicate synaptic pathways{; those with open}-circle heads indicate excitation/additive pressure, solid square heads indicate inhibitory/subtractive pressure and diamond heads indicate modulation/multiplicative \textcolor{black}{pressure}. 
        {Dashed lines (with solid triangle heads)} indicate non-synaptic pathways ({where} no transformation {is} applied).
        } 
        \label{fig:csdp_dynamics}
\end{figure}

In Equation 5, 
we simulate an adaptive, non-negative spiking threshold $v^\ell_{thr} \in [0,\infty)$ which, at time $t$, either results in the addition or subtraction of a small value -- scaled by $\lambda_v$, which is a constant typically set to a number such as $0.001$ -- from a current threshold value based on how many spikes were recorded at time $t$ for layer $\ell$. This represents a simple homeostatic constraint on short-term plasticity, encouraging each layer to emit as few spikes as possible at any step in time. Finally, notice that $\mathbf{s}^0(t)$ is the binary spike representation of the sensory input $\mathbf{x}$ at time $t${. A} spike is created from this pattern vector, e.g., an image, by sampling the normalized pixel values of $\mathbf{x}$ (each {value/element} of $\mathbf{x}$ is divided by a scalar, such as the maximum pixel value $255$), treating each {value} as Bernoulli probability. {In this study,} $\mathbf{s}_y(t)$ is the target class spike train associated with $\mathbf{x}${. If} a label is available, then $\mathbf{s}_y(t) = \mathbf{y}$ (at each time step, it is clamped to the sparse label during training), otherwise $\mathbf{s}_y(t) = \emptyset$ {(since $\mathbf{y} = \emptyset$)}.

Notice that the key property of the recurrent spiking system above is that, by design, it is layer-wise parallelizable and therefore naturally not forward-locked \cite{jaderberg2017decoupled}{. In essence,} each layer's spike emissions can be computed in parallel {to} the other layers. In addition, as will be seen later, the system is not update/backward-locked -- changes in synaptic strengths for any layer can be computed in parallel of the others -- due to the fact that our plasticity scheme operates with information that is local in both time and space. This stands in contrast to many (deep) SNN designs today, where each layer of spiking NPUs {depends} on the previous spike activities of the layer that comes before them. Other models that are not forward-locked, such as spiking predictive coding schemes \cite{rao2004hierarchical,ororbia2019spiking,ndri2023}, generally rely on introducing additional neuronal circuitry in the form of feedback cycles which increases the complexity of the neural computation that {underlies} inference. Our model ensures an architectural parallelism by only enforcing the leaky integrator spike activities to depend upon recently computed spatially-local activities, which is biologically more plausible as well as efficient and cheap to compute. 

\begin{figure}[!t]
    \begin{center}
    \includegraphics[width=0.75\textwidth]{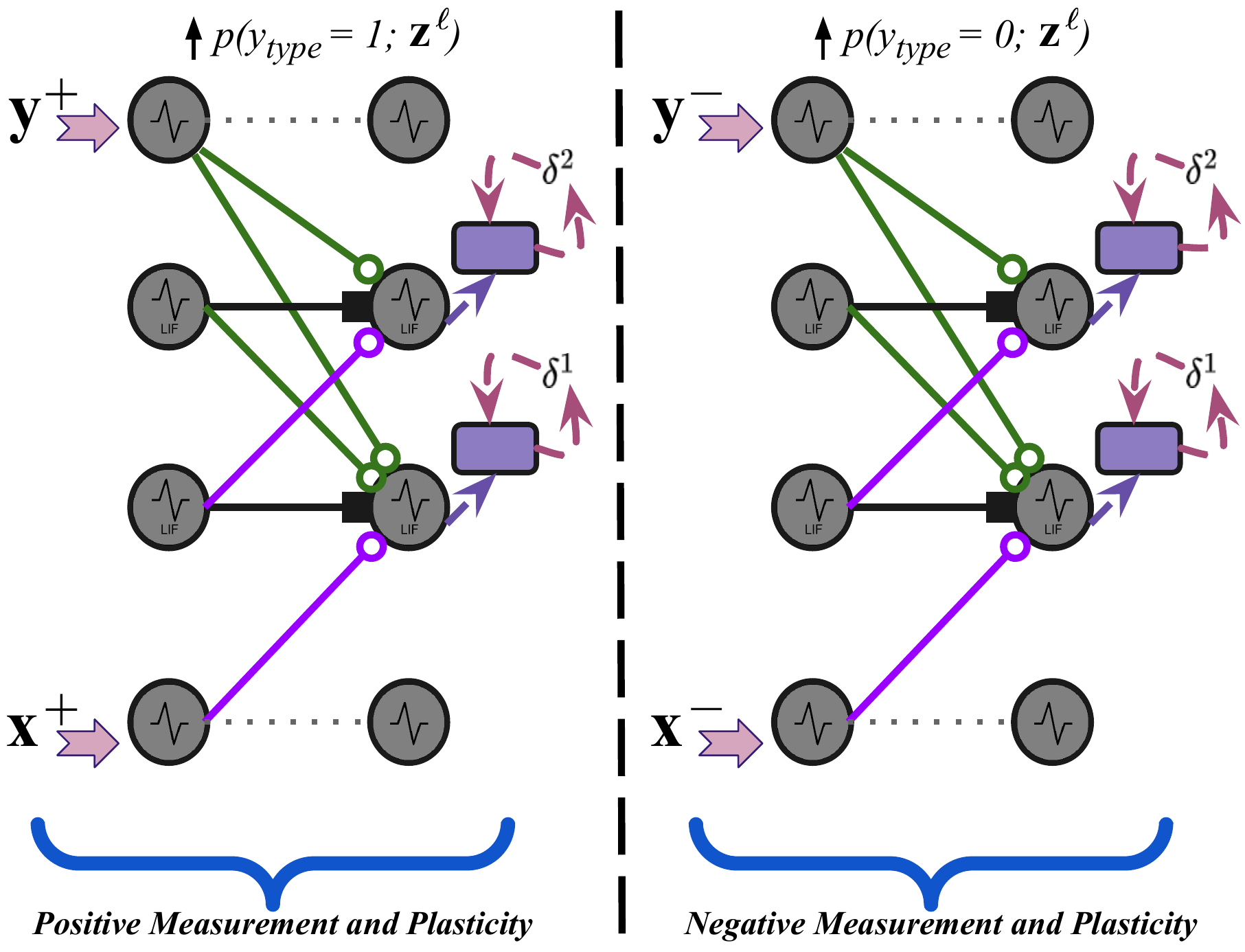}
    \end{center}
    \vspace{-0.2cm}
    \caption{\textbf{Positive and negative modes of CSDP plasticity in a recurrent SNN.} 
    Shown are the two parallel `modes' of plasticity that our recurrent spiking circuit undergoes -- a positive measurement mode is induced when a sensory input and its corresponding (if available) {target class} are used to drive inference and plasticity (this entails neural layers increasing the probability values assigned to a sensory pattern), and a negative measurement mode is induced when a negative / out-of-distribution sample and its corresponding negative {target class} are used to drive the same underlying neural calculations (this entails neural layers lowering probabilities that they ultimately assign). {Solid square arrows denote inhibitory pressure whereas open circle arrows denote excitatory pressure.  
    Colors for different synaptic pathways denoted by solid lines, all of which are plastic, are provided to visually distinguish between top-down class modulating synapses (green), bottom-up sensory driving synapses (purple), and lateral synapses (black).}
    }
    \label{fig:csdp_modes}
\end{figure}

\subsection*{Plasticity Dynamics: Contrastive-Signal-Dependent Plasticity}

Key to the long-term plasticity process that we propose in this study is the trace variable. Practically, this will mean that an additional {component}, responsible for maintaining what is called an `activation trace', is introduced to each neuronal cell of the recurrent spiking system. As a result, for each layer, a trace is dynamically updated in the following manner:
{
\begin{equation}
    z^\ell_i(t+\Delta t) = z^\ell_i(t) + (\Delta t/\tau_{tr}) \Big( -z^\ell_i(t) + \gamma s^\ell_i(t + \Delta t) \Big) 
\end{equation}
}
where, in this work, the trace time constant was set to $\tau_{tr} = 3$ ms and $\gamma = 0.05$. 
Importantly, an activation trace smooths out and filters the sparse spike train signals generated by each layer of NPUs while still being biologically-plausible. Mechanistically, a trace offers a dynamic rate-coded equivalent value that, in our model, would likely be biophysically instantiated in neuronal cells in the form of concentrations of internal calcium ions \cite{oreilly2000computational}.
Desirably, this will allow our adaptation rule to take place using (filtered) spike information, bringing it closer to a form of STDP-like adjustment \cite{bi1998synaptic}, rather than directly operating on voltage or electrical current values, as in related efforts in developing credit assignment processes for SNNs \cite{xie2017efficient,kohan2023signal}.

Given the aforementioned activation trace, the objective that each layer of our spiking system will try to optimize, at each step of simulated time, may next be specified. Specifically, in our plasticity process, \emph{contrastive-signal-dependent plasticity} (CSDP), we propose adjusting SNN synaptic strengths by integrating the `goodness principle' inherent to forward-forward learning \cite{hinton2022forward,ororbia2022predictive} into the dynamics of spiking neurons. This, in turn, will mean that any layer of NPUs will work to raise the probability they (collectively) assign to the incoming pre-synaptic messages that they receive if the sensory input comes from the environment (it is `positive'){. Conversely, NPUs} will strive to lower the probability that they assign if the input is fake/non-sensical (it is `negative' or out-of-distribution); see Fig. 3. 
As a result, the (local) {contrastive function} takes the following form:
\begin{equation}
    \mathcal{C}(\mathbf{z}^\ell(t),y_{type}) = -\Big( \overbrace{y_{type} \log p\big( y_{type}=1; \mathbf{z}^\ell(t) \big)}^{\text{Positive Term}} + 
    \overbrace{(1 - y^\ell_{type}) \log p\big( y_{type}=0; \mathbf{z}^\ell(t) \big)}^{\text{Negative Term}} \Big), \label{eqn:local_loss}
\end{equation}
where the goodness probability $p(y_{type}=1; \mathbf{z}^\ell)$ that will be computed by {a group} of spiking neurons in any layer $\ell$ can be directly defined as:
\begin{equation}
    p\big(y_{type}=1; \mathbf{z}^\ell(t)\big) = 1/\Big( 1 + \exp \big( -(\Sigma^{J_\ell}_{k=1} (z^\ell_k(t))^2 - \theta_z ). \big) \Big) 
\end{equation}
Note that $y_{type} \in \{0,1\}$ is an integer which denotes what `type' of sensory input a sample of incoming (pre-synaptic) activity signals feeding into (post-synaptic) layer $\ell$ originate from, i.e., $1$ indicates that such signals originate from or are caused by positive (in-distribution) sensory input while $0$ indicates that they come from negative (out-of-distribution) input. $\theta_z$ is a fixed threshold {against which the square of the trace values are compared} {and is set to the value $\theta_z = 10$}. 
As a result of the local contrastive {function} (Equation 7), 
the adjustments made to synaptic bundles for any layer become simple, efficient modulated Hebbian-like rules {(see the supplementary text for derivations and connections made to STDP)}. For instance, with respect to the synapses $\mathbf{W}^\ell$ connecting the layer below $\ell-1$ to layer $\ell$, the resultant update would be:
\begin{equation}
    \delta^\ell_i(t) =  \partial \mathcal{C}(\mathbf{z}^\ell(t), y_{type}) \Big/ \partial z^\ell_i(t)  \\  
\end{equation}
\begin{equation}  
    \Delta W^\ell_{ij} = \Big( R_m \delta^\ell_i(t) s^{\ell-1}_j(t-\Delta t) \Big) + \lambda_d \Big( s^\ell_i(t) \big(1 - s^{\ell-1}_j(t - \Delta t)\big) \Big)
\end{equation}
where $\lambda_d$ is the synaptic decay factor {and $\delta^\ell(t) \in \mathcal{R}^{J_\ell \times 1}$ (a vector containing one modulatory signal per neuron). See} Materials and Methods for the updates {with respect to} all possible sets of synapses. 
Ultimately, the updates to all four possible synaptic bundles that transmit information to and drive the dynamics of layer $\ell$ depend on the easily computed partial derivative(s)  $\delta^\ell(t)$ of the local contrastive function $\mathcal{C}(\mathbf{z}^\ell(t), y_{type})$. We postulate {that the vector signal} $\delta^\ell(t)$, scaled by the resistance, would be biologically implemented either in the form of a (neurotransmitter) chemical signal or, alternatively, as a separate population of NPUs that synthesize this signal, much akin to the error neurons that characterize predictive processing \cite{clark2015surfing}. 
Furthermore, the above synaptic plasticity equations could be viewed as a special kind of multi-factor Hebbian rule where the modulator term is deposited from another locally embedded neural {component} (within any one NPU) that biochemically signals the contrastive {function} $\mathcal{C}(\mathbf{z}^\ell(t), y_{type})$, as opposed to the typical temporal (reward) difference error found in three-factor Hebbian/STDP update schemes. See Fig. 2 
for a depiction of the mechanics that underpin CSDP. Our local contrastive modulatory factor desirably stands on the more recent empirical evidence found in support of self-supervised learning in neurobiological systems \cite{konkle2022self,nayebi2023neural}. The two general `modes' of plasticity that our system undergoes are depicted in Fig. 3
{. Specifically,} if an in-distribution sensory pattern (and possibly any corresponding {class target vector}) is fed in, then the neuronal system tries to increase the probability assigned to the pattern whereas, for an out-of-distribution pattern, it tries to decrease this probability. 

Holistically, our neural dynamics and plasticity process {could be treated as operating under a total goodness contrastive} functional (function of functions) over the entire neural circuit. Specifically, this objective {entails viewing our neural circuit as} optimizing, {iteratively} across a span of time, {the following}: 
\begin{equation}
    \mathcal{E}(\theta) = \sum^T_{t=1} \mathcal{F}(t, \Theta) = \sum^T_{t=1} \sum^L_{\ell=1} \mathcal{C}\Big(\mathbf{z}^\ell(t), y_{type}\Big)
\end{equation}
where $\Theta$ is a construct that stores every single bundle of synapses (see Materials and Methods) and $\mathcal{C}(\mathbf{z}^\ell(t), y_{type})$ is the local goodness function as defined in Equation 7 {(from which synaptic updates for synapses related to layer $\ell$ come from; see the supplementary text for details)}. 
{Total goodness is effectively} an aggregation of all of the goodness measurements produced by every layer of the neural circuit. 
In effect, a full recurrent spiking circuit, under our framework, is optimizing {, at time $t$,} its distributed representations of sensory inputs with respect to a sequence loss $\mathcal{E}(\Theta)$ which sums, over all time-steps, the total system-level goodness values $\mathcal{F}(t, \Theta)$. 

\subsection*{Synthesizing Negative Patterns for Supervised and Unsupervised CSDP} 
Given that CSDP inherently centers around self-supervised contrastive adaptation, a process for producing negative or out-of-distribution data patterns to contrast with encountered sensory patterns must be provided. {Note that, based on the dynamics presented above, two general variants of CSDP are possible: a supervised variant if the class-modulating synapses $\mathbf{B}^\ell$ are used and an unsupervised variant if they are not. Depending on which variant is employed, a different scheme \textcolor{black}{is used} for generating out-of-distribution/negative data points that \textcolor{black}{further} operated on-the-fly, i.e., only utilized information/statistics within a current batch or small buffer of pooled data samples. }

In the case of the supervised model, we adapted the scheme used in \cite{hinton2022forward,ororbia2022predictive}{. For} each $(\mathbf{y}, \mathbf{x})$ in a mini-batch/pool of patterns, we would produce a negative batch $(\mathbf{y}^-, \mathbf{x}^-)$ by duplicating the original images and, for each cloned (original/positive) pattern, we then created a corresponding `incorrect' paired label by randomly sampling a class index other than the correct one (index $c$) found within $\mathbf{y}$. Formally, if $c \in \{1,...C\}$ is the correct class index of $\mathbf{y}$, then a negative label, with incorrect class index $q \neq c$, is produced via: $\mathbf{y}^- = \mathbf{1}_q$ where $q \sim \{1,...,c-1,c+1,...,C\}$. 
For the unsupervised variant, where $\mathbf{y} = \emptyset$, we took inspiration from \cite{zhang2018mixup} and designed an on-the-fly process that applied the following steps to a current mini-batch of patterns: 
\textbf{1)} for each pattern $\mathbf{x}(i)$, we would select one other different pattern $\mathbf{x}(j)$, $j \neq i$, within the batch, 
\textbf{2)} apply a random rotation to $\mathbf{x}(j)$ (by sampling a rotation value, in radians, from the range $\big(\pi/4, (7\pi)/4\big)$ 
to create $\mathbf{r}(j)$, and finally, 
\textbf{3)} produce a negative pattern via the convex combination $\mathbf{x}^-(i) = \eta \mathbf{x}(i) + (1 - \eta) \mathbf{r}(j)$ with $\eta = 0.55$ ($\eta$ interpolates between the original and non-original pattern). 

\subsection*{Task-Specific Generalization}
The full neural circuit described above is effectively a spiking representation learning system{. In} other words, {it} does not engage in any particular behavioral task. 
To extend our circuit so that it exhibits {task-specific} functionality, we investigated modifications that facilitated sensory pattern reconstruction and {label} classification. Specifically, to promote reconstruction, we integrated, in each layer $\ell$, a small bundle of extra synapses that made local predictions of the spike activity of the layer immediately below $\ell-1$. {These local prediction units also follow the same dynamics as the LIF layers described earlier, e.g., Equations 2-4, but instead adapt with an error-driven Hebbian rules.} 
To facilitate classification, we integrated a small spiking sub-circuit that directly wired the spike emissions of each (non-input) layer to a single output prediction of the {target class} spike train $\mathbf{s}_y(t)$ -- this speeds up the test-time model inference for the fixed memory cost of several extra synaptic matrices. \textcolor{black}{The output units that predicted} the class spike train again \textcolor{black}{follow} the same LIF dynamics as described before but \textcolor{black}{their specific synapses}, like \textcolor{black}{those of} the local reconstruction units, \textcolor{black}{are} adapted with error-driven Hebbian rules. 
See Materials and Methods for {the formal details related to} our framework's biophysical generative and classification dynamics and plasticity processes. 

\subsection*{Classification Performance} 

\begin{table}[!t]
\begin{center}
\begin{tabular}{c | c | c| c | c} 
 \hline
  & \textbf{MNIST} & \textbf{K-MNIST} & \\
  & \textbf{ACC} (\%) & \textbf{ACC} (\%) & {\textbf{NoPS}} & \textcolor{black}{\textbf{NoS}}\\
 \hline\hline
 BP-FNN (Imp.) & $98.70 \pm 0.02$ & $93.66 \pm 0.07$ & {$54,572,010$} & \textcolor{black}{$54,572,010$}\\
 \hline 
 Spiking-RBM \cite{merolla2011digital} & $89.00$ & -- & {$250,368$} & \textcolor{black}{$250,368$}\\ 
 \hline
 {ML H-SNN \cite{beyeler2013categorization}} & {$91.64$} &  -- & {$3,136$} & \textcolor{black}{$3,136$}\\
 \hline
 SNN-LM \cite{hazan2018unsupervised} & $94.07$ & -- & {$1,254,400$} & \textcolor{black}{$1,254,400$} \\ 
 \hline
 Power-Law STDP \cite{diehl2015unsupervised} & $95.00$ & -- & {$5,017,600$} & \textcolor{black}{$86,937,600$}\\ 
 \hline 
 STDP-SNN 
 \cite{hao2020biologically} (Imp.) & $96.05 \pm 0.15$ & $73.82 \pm 0.18$ & {$7,840,000$} & \textcolor{black}{$207,840,000$}\\ 
 \hline
 DRTP \cite{frenkel2019learning} (Imp.) & $93.55 \pm 0.17$ & $78.68 \pm 0.13$ & {$54,572,010$} & \textcolor{black}{$54,712,010$} \\ 
 \hline
 BFA \cite{samadi2017deep} (Imp.) & $96.65 \pm 0.10$ & $88.75 \pm 0.05$ & {$54,572,010$} & \textcolor{black}{$54,712,010$}\\ 
 \hline
 L2-SigProp \cite{kohan2023signal} (Imp.) & $88.58 \pm 0.02$ & $70.94 \pm 0.04$ & {$201,726,010$} & \textcolor{black}{$201,726,010$}\\ 
 \hline 
 Loc-Pred \cite{mostafa2018deep} (Imp.) & $93.76 \pm 0.03$ & $75.18 \pm 0.20$ & {$54,782,020$} & \textcolor{black}{$54,782,020$} \\ 
 \hline 
 CSDP, Unsup (Ours; \textit{Sm}) & $95.12 \pm 0.07$ & $86.55 \pm 0.04$ & {$7,791,000$} & \textcolor{black}{$7,791,000$}\\ 
 \hline 
 CSDP, Unsup (Ours; \textit{Lg}) & $95.96 \pm 0.05$ & $89.43 \pm 0.16$ & {$39,980,000$} & \textcolor{black}{$39,980,000$}\\ 
 \hline
 CSDP, Sup (Ours; \textit{Sm}) & $97.02 \pm 0.04$ & $88.49 \pm 0.12$ & {$7,815,500$} & \textcolor{black}{$7,815,500$}\\
 \hline
 CSDP, Sup (Ours; \textit{Lg}) & $97.58 \pm 0.05$ & $91.53 \pm 0.15$ & {$40,040,000$} & \textcolor{black}{$40,040,000$}\\ 
 \hline
\end{tabular}
\caption{ \textbf{Generalization of SNNs trained under different bioplausible schemes.} 
Measurements of generalization accuracy (ACC, in terms of \%) of spiking networks trained with different credit assignment processes (means and standard deviations reported for $10$ trials). ``Imp.'' denotes implementation \textcolor{black}{whereas the label `Sm' indicates a small CSDP model and the label `Lg' indicates a large CSDP model}. BP-FFN is the rate-coded comparison model (a backprop-trained feedforward neural network). {Number of plastic synapses (NoPS) \textcolor{black}{and total number of synapses (NoS) are} also reported for each model (\textcolor{black}{these counts include} all classification parameters). Note that \textcolor{black}{training the} spiking RBM model \textcolor{black}{requires} shrinking the MNIST inputs to a shape of $22\times22$ pixels \cite{merolla2011digital}.} \textcolor{black}{Note that both the unsupervised (`Unsup') and supervised (`Sup') CSDP models were each equipped with separate, local classification synapses that were trained alongside and simultaneously with the rest of their parameters, allowing both model variants to readily make class predictions at any point in time (see the section ``Spike-Driven Classification and its Fast Approximation'' for details).}}
\label{table:benchmarks}
\vspace{-0.5cm}
\end{center}
\end{table}

\begin{table}[!t]
\begin{center}
\begin{tabular}{c | l | c | c | c | c} 
 \hline
 &  & \multicolumn{2}{|c|}{\textbf{Supervised}}  &  \multicolumn{2}{|c}{\textbf{Unsupervised}}\\
 \hline
  & \textbf{Batch Size} & \textbf{ACC} (\%) & \textbf{BCE} (nats) &  \textbf{ACC} (\%) & \textbf{BCE} (nats)  \\
 \hline\hline 
 \parbox[t]{2mm}{\multirow{5}{*}{\rotatebox[origin=c]{90}{\footnotesize MNIST}}} & $B = 2$ & {$95.19 \pm 0.13$} & {$143.60 \pm 0.24$} & {$94.87 \pm 0.07$} & {$131.74 \pm 1.21$} \\ 
  & $B = 20$ & {$96.55 \pm 0.11$} & {$138.28 \pm 0.40$} & {$95.18 \pm 0.22$} & {$127.97 \pm 0.33$}\\ 
  & $B = 50$ & {$96.84 \pm 0.14$} & {$138.02 \pm 0.71$} & {$95.44 \pm 0.25$} & {$128.44 \pm 0.22$} \\ 
  & $B = 100$ & {$97.05 \pm 0.08$} &  {$135.11 \pm 1.29$} & {$95.62 \pm 0.11$} & {$126.36 \pm 1.91$}\\ 
  & $B = 200$ & {$97.16 \pm 0.07$} & {$135.84 \pm 0.43$} & {$95.75 \pm 0.28$} & 
  {$126.61 \pm 0.79$} \\ 
  & $B = 500$ & {$97.58 \pm 0.05$} & {$134.23 \pm 0.59$} & {$95.96 \pm 0.05$} & 
  {$126.09  \pm  0.83$} \\ 
 \hline 
 \parbox[t]{2mm}{\multirow{5}{*}{\rotatebox[origin=c]{90}{\footnotesize K-MNIST}}} & $B = 2$ & {$86.35 \pm 0.22$} & {$344.25 \pm 0.84$} & {$85.92 \pm 0.34$} & {$286.46 \pm 1.36$}\\
  & $B = 20$ & {$87.02 \pm 0.31$} & {$331.93 \pm 0.92$} & {$86.81 \pm 0.16$} & {$286.30 \pm 0.99$}\\
  & $B = 50$ & {$88.30 \pm 0.17$} & {$324.14 \pm 1.22$} & {$87.81 \pm 0.21$} & {$287.21 \pm 0.92$}\\ 
  & $B = 100$ & {$90.00 \pm 0.13$} & {$315.13 \pm 0.55$} & {$88.76 \pm 0.16$} & {$282.64 \pm 0.61$}\\ 
  & $B = 200$ & {$90.44 \pm 0.12$} & {$313.57 \pm 1.32$} & {$88.51 \pm 0.13$} & {$285.76 \pm 2.12$}\\ 
  & $B = 500$ & {$91.53 \pm 0.15$} & {$303.36  \pm  0.96$} & {$89.43 \pm 0.16$} & 
  {$284.03  \pm  0.81$} \\ 
 \hline 
\end{tabular}
\caption{ \textbf{CSDP generalization performance across batch sizes.} Measurements of generalization accuracy (ACC, in terms of \%; higher is better) and reconstruction binary cross entropy (BCE, in terms of nats; lower is better) on the development sets of MNIST and K-MNIST (means and standard deviations reported for $10$ trials). The SNN model examined for this experiment consisted of $5000$ LIFs in the first hidden layer and $1000$ LIFs in the second hidden layer. Each batch-size ($B$) variant model was adapted over a stream \textcolor{black}{of $1.5 \times 10^6$ samples ($30$} epochs worth of pattern data) for MNIST and K-MNIST. 
\textcolor{black}{Note that both the unsupervised (`Unsup') and supervised (`Sup') CSDP models were each equipped with separate, local classification synapses that were trained alongside and simultaneously with the rest of their parameters, allowing both model variants to readily make class predictions at any point in time (see the section ``Spike-Driven Classification and its Fast Approximation'' for details).}
}
\label{table:batch_results}
\vspace{-0.5cm}
\end{center}
\end{table}


To evaluate the efficacy of our recurrent spiking circuitry and the proposed CSDP scheme, we conducted several experimental simulations in visual symbol recognition, using the MNIST and Kuzushiji-MNIST (K-MNIST) datasets. We simulate both unsupervised and supervised variants of the CSDP modeling framework -- for unsupervised models we use the label `CSDP, Unsup' whereas for supervised 
models, we use the label `CSDP, Sup' -- and compare with several previously reported STDP-adapted SNN results as well as several implemented spiking network biophysical credit assignment baselines. The particular baselines that we implemented and study include: 
\textbf{1)} a spiking network classifier trained with direct random target propagation (DTRP) \cite{frenkel2019learning}, 
\textbf{2)} a spiking network trained with broadcast feedback alignment (BFA) \cite{samadi2017deep}, 
\textbf{3)} a spiking network trained with a simplified variant of signal propagation \cite{kohan2018error,kohan2023signal} using a voltage-based rule and a local cost based on the Euclidean (L2) distance function (L2-SigProp),  
\textbf{4)} a spiking network trained using a generalization of local classifiers/predictors, where each layer-wise classifier, instead of being held fixed, is learned jointly with the overall neural system \cite{mostafa2018deep,zhao2023cascaded} (Loc-Pred)\textcolor{black}{, and 5) an STDP-trained spiking network \cite{hao2020biologically} (STDP-SNN), adapted to operate under this work's experimental settings (we utilized the simultaneously-trained model architecture of \cite{hao2020biologically}, which contained one very wide hidden layer of $10,000$ LIFs, since the authors of  \cite{hao2020biologically} demonstrated that STDP processes struggle to train networks with more than one hidden layer, offering no further benefit beyond a single-layer formulation). 
}
Furthermore, we provide a rate-coded baseline model, i.e., one that does not operate with spike trains but instead with rate-coded values, represented by a tuned feedforward neural network trained by backpropagation of errors (BP-FNN). Notably, SNNs learned via BFA, DRTP, or Loc-Pred require additional neural circuitry in order to create either feedback loops or layer-wise local predictors\textcolor{black}{, whereas L2-SigProp and the STDP-SNN do not}.

The network models simulated under each learning algorithm were designed to have two latent/hidden layers, with each consisting of up to a maximum of $7,000$ leaky integrate-and-fire (LIF) neurons \textcolor{black}{(except for STDP, which has one layer of two laterally-wired groups of $10,000$ LIFs)}. \textcolor{black}{We further study and report, in Table 1,  
for both supervised and unsupervised cases of CSDP, the performance of a `small' CSDP model (labeled as `Sm'), which contained $2250$ LIFs in the first layer and $200$ LIFs in the second, as well as that of a `large' CSDP model (labeled as `Lg'), which contained $5000$ LIFs in the first layer and $1000$ LIFs in the second. The smaller model, as a result, is made up of about only $7.8$ million synapses (thus, it is comparable to the amount of plastic synapses used in the STDP-SNN baseline) whereas the larger model is made up of about $40$ million synapses (thus, it is roughly comparable to the amount of plastic synapses used in the bigger feedback-driven baselines).
} 
Synaptic connection efficacies \textcolor{black}{were} randomly initialized from a {uniform distribution $\sim \mathcal{U}(-1,1)$} and truncated to the range of $[-1,1]$ (this truncation was enforced throughout training), except for any lateral synapses, which were initialized via $\sim \mathcal{U}(0,1)$ and always truncated to the values in the range of $[0,1]$. All SNN models further employed a spiking classifier of the same design -- aggregating the spike vector outputs across all layers -- as the one proposed for our CSDP SNN system in order to ensure a fair comparison. In general, we found these hidden-to-output synapses improved generalization performance across the board. All models/baselines were trained for a fixed $30$ epochs for each database.
 
Since all of the neurobiologically-plausible credit assignment algorithms, including our own, could support mini-batch calculations, we trained all models using mini-batches of $500$ patterns, randomly sampled without replacement from a training dataset, to speed up simulation. {Note that, for CSDP, only one negative sample is created on-the-fly for each positive sensory sample and, for simulation efficiency, all negative samples are appended to the mini-batch of original (sensory) samples.} 
Furthermore, the Adam adaptive learning rate \cite{kingma2014adam} (with step size $\eta = 0.002$) was leveraged to physically adjust synaptic strength values utilizing the updates provided by any one of the simulated algorithms. Each data point/batch 
was presented to all spiking networks 
for a stimulus window of $T = 90-150$ ms ($\Delta t = 3$ ms). 
Finally, note that all spiking networks employed the same adaptive threshold update scheme that the CSDP system used -- it was observed that this mechanism improved training stability in all cases. 

To quantitatively compare models in terms of behavior (i.e., with respect to categorizing the processed visual symbols), we measured, and report in Table 1, 
the generalization {accuracy on unseen (test) samples.} 
{Note that accuracy was calculated in the following manner:} ${\mbox{acc} = (1/N) \sum^N_{n=1} (\argmax_{c\in C} \mathbf{Y}_n \equiv \argmax_{c\in C} \mathbf{\bar{Y}}_n)}$ {where} the matrix $\mathbf{Y} \in \{0,1\}^{N \times C}$ contains all of the one-hot encoded labels \textcolor{black}{($\mathbf{Y}_n$ retrieves/extracts the $n$th row vector of $\mathbf{Y}$)} while the matrix $\mathbf{\bar{Y}} \in \mathcal{R}^{N \times C}$ contains all of the collected model predicted class probability vectors (see Materials and Methods for how these probabilities were estimated from the spiking networks). Generalization {accuracy} values in Table 1 are reported in terms of the mean and standard deviation, i.e., $\mu \pm \sigma$. These statistics were calculated across $10$ experimental simulation trials (the random number generator for each trial was seeded with a unique integer). \textcolor{black}{Furthermore, to contextualize the results in terms of biophysical model parameter complexity, we measure and report the number of plastic synapses (NoPS) of each model, for both implemented and prior reported results, as well as the total number of synapses (NoS).}

\begin{figure}[!t]
     \centering
     \includegraphics[width=0.95\textwidth]{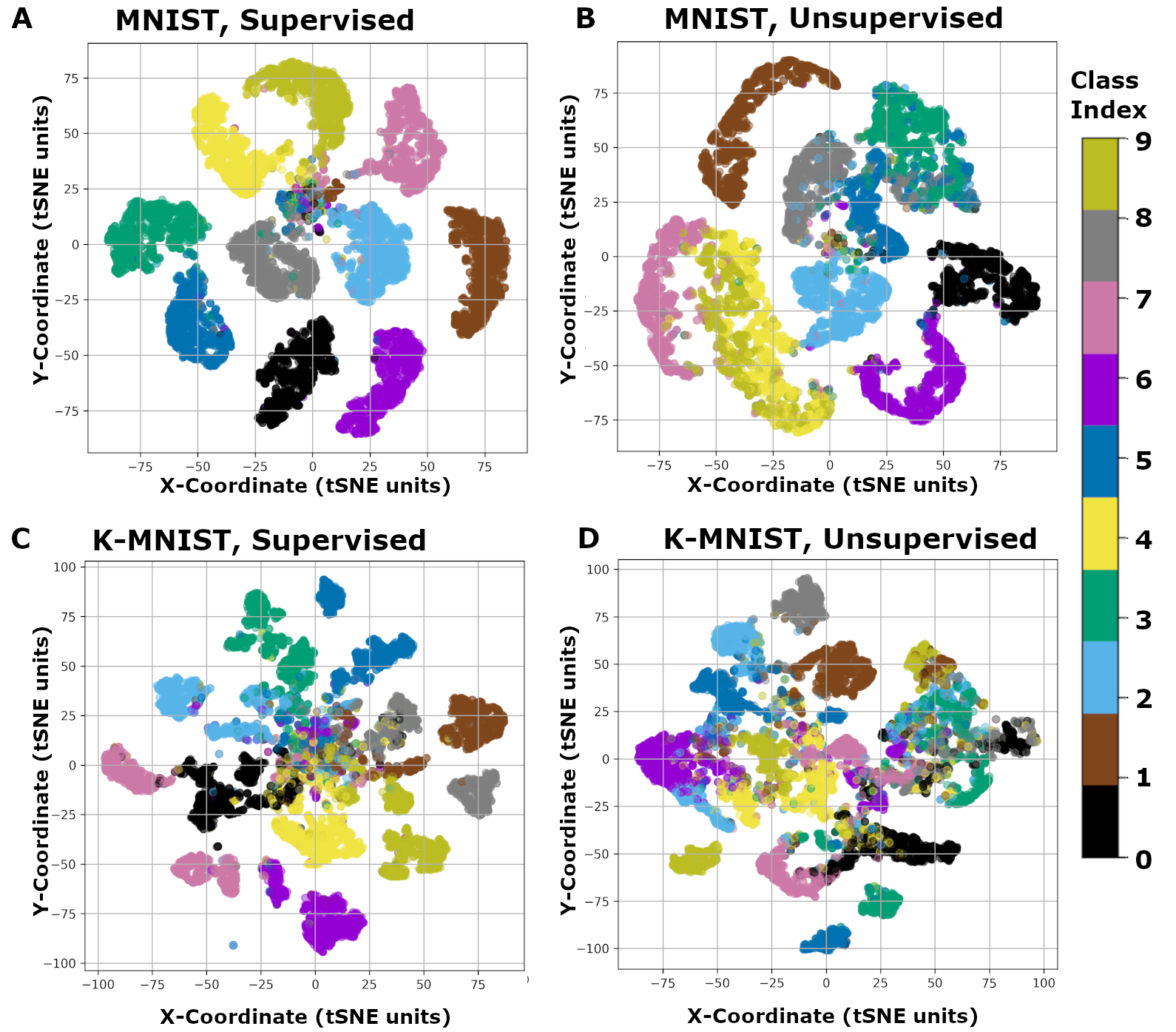}
        \caption{ \textbf{Latent activity patterns acquired by CSDP-trained SNNs.} 
        t-SNE visualizations of the latent space induced by recurrent spiking networks learned with the CSDP process. {Note that t-SNE coordinate units are technically dimensionless units and we thus denote them as ``tSNE units\textcolor{black}{''}).} Rate codes are visualized for: 
        (\textbf{A}) `CSDP, Sup' (supervised  CSDP-trained SNN) on MNIST, 
        (\textbf{B}) `CSDP, Unsup' (unsupervised CSDP-trained SNN) on MNIST, 
        (\textbf{C}) `CSDP, Sup' on K-MNIST, and 
        (\textbf{D}) `CSDP, Unsup' on K-MNIST. 
        }
        \label{fig:latent_viz}
        \vspace{-0.5cm}
\end{figure}

In Table 1, 
we present the results of our experimental simulations. Notice that, like many biophysical spiking networks, although we do not exactly match the performance of backprop-based feedforward networks (BP-FNNs), our generalization error comes promisingly close. This is despite the fact that every layer of our model is a group of spiking neurons, making inference much noiser than in models that operate with clean continuous rate-coded activities. Furthermore, our CSDP models, particularly the supervised  variant `CSDP, Sup', comes the closest to matching the BP-FNN rate-coded baseline as compared to other SNN credit assignment procedures. The unsupervised CSDP model only does a bit worse than the supervised one yet effectively operates \textcolor{black}{on par with the STDP-SNN and} nearly on par with the BFA-adapted SNN on MNIST, further notably outperforming \textcolor{black}{(both)} on K-MNIST. We remark that the BFA SNN offers a competitive baseline -- feedback synaptic pathways empirically were found to result in improved overall performance compared to other schemes such as Loc-Pred and L2-SigProp -- and, thus, it is promising to observe that CSDP, which does not require any additional feedback connectivity, still effectively outperforms it in terms of generalization ability on test image samples. \textcolor{black}{The STDP-SNN offers a competitive non-feedback based baseline on MNIST, although it is a single-layer model (and struggles to work well on deeper architectures, as discussed in \cite{hao2020biologically}, obtaining only $83.75$\% on MNIST when applied to a two-hidden layer SNN), yet does not perform well on the harder K-MNIST database. However, on both datasets, the supervised variant of CSDP (both small `Sm' and `Lg' models) offers better generalization than the STDP-SNN (even compared to the $96.73$\% maximal MNIST score reported in \cite{hao2020biologically}, which required a greedy, layer-by-layer training heuristic scheme), even when CSDP is restricted to using nearly the same number of plastic parameters as the STDP-SNN, i.e., $7$-$8$ million plastic synapses. This is despite the fact that the STDP-SNN is able to further enjoy the benefit of using a much larger pool of non-plastic synapses, as revealed by its NoS measurement.} 

{
In Table 2
, we conducted one more experiment to investigate the effect that batch size had on CSDP's ability to yield effective SNN models, both in terms of classification and sensory input reconstruction ability. 
\textcolor{black}{We trained the model under different batch size conditions, ranging from size $2$ to $500$. }
Observe, in Table 2, that there is a small dependency between model performance and batch size (a known issue with many forms of self-supervised learning in general). Decreasing the batch size from $500$ towards $2$ leads to, in the worst case, a drop in accuracy of \textcolor{black}{a bit more than $2$\% on} MNIST and more \textcolor{black}{than $5$\%} on K-MNIST for the supervised CSDP model while there was a degradation of about \textcolor{black}{$1$\% on} MNIST and \textcolor{black}{$3.5$\% on} K-MNIST for the unsupervised variant (reconstruction cross-entropy also went up/degraded by several nats).
}

\subsection*{Visualizing CSDP-SNN Latent Distributions}

In Fig. 4, 
we next examined the latent space induced by recurrent spiking circuits trained with CSDP. 
Specifically, we collected the neural `codes' produced by the top-most hidden layer of each model by converting each temporal spike train produced (in response to a single pattern) to {a} single rate-code vector $\mathbf{c}$. 
For each data point, we calculate an approximate real-valued, rate-code vector from the circuit's top layer spike train, i.e., $\{\mathbf{s}^L(t)\}^T_{t=1}$, as follows:
\begin{equation}
    \mathbf{c} = (\gamma_c/T) \sum^T_t \mathbf{s}^L(t) 
\end{equation}
with $\gamma_c = 1$. After feeding each data point $\mathbf{x}$ (further omitting its {target class input} $\mathbf{y}$ if one is available) in the test-set to each model and collecting an approximate top-layer rate-code vector $\mathbf{c}$, we visualized the emergent islands of rate codes using t-Distributed Stochastic Neighbor Embedding (t-SNE) \cite{van2008visualizing}. 
Notice how, for both MNIST (Fig. 4 A {and B}) 
and K-MNIST (Fig. 4 {C and} D), 
clusters related to each category form, qualitatively demonstrating why CSDP-adapted SNNs are able to classify unseen test patterns effectively. This is particularly impressive for K-MNIST, which is arguably the more difficult of the two datasets, which means that CSDP adaptation is capable of extracting class-centric information from more complicated patterns. Finally, observe that the clusters are less separated/distinct in the case of `CSDP, Unsup', i.e., Fig. 4 B and D, 
as opposed to `CSDP, Sup' (which is due to the lack of {target class} mediation).

\subsection*{Pattern Reconstruction and Receptive Fields} 
In Fig. 5, 
we visualize the pattern reconstruction ability of a CSDP-adapted spiking model (focusing on the `CSDP, Sup' variant) on randomly sampled, without replacement, image patterns from the MNIST and K-MNIST databases. Desirably, we see that the CSDP model is able to produce reasonable reconstructions of the image patterns,  demonstrating {that it is able to encode, in its local generative synapses, information needed to decode the sparse spike train representations stored in its internal layers back into sensory input space.}

We finally investigate the representation capability of our spiking system in the context of image patches. Specifically, we extracted patches of $8 \times 8$ and $12 \times 12$ pixels from a subset of the MNIST database ($1000$ image patterns, $10$ from each class) and simulated a CSDP-learned model with two layers of $2000$ leaky integrator NPUs each. The learned receptive fields, for each of the two patch settings, of the bottom-most layer of the trained model, i.e., layer $\ell = 1$, are shown in Fig. 6. 
Receptive fields in the bottom layer were extracted for visualization by randomly sampling, without replacement, $100$ slices of the synaptic matrix $\mathbf{W}^1$. As observed in Fig. 6, 
the receptive fields extract basic patterns at different resolutions{. In} the smaller $8 \times 8$ patch model, we see simpler patterns emerge, such as edges or stroke pieces (``strokelets'') of different orientations. In the larger $12 \times 12$ patch model, we observe portions of patterns that appear as ``mini-templates'' that could be used to compose complete digit symbols.

\begin{figure}[!t]
     \centering
     \includegraphics[width=0.875\textwidth]{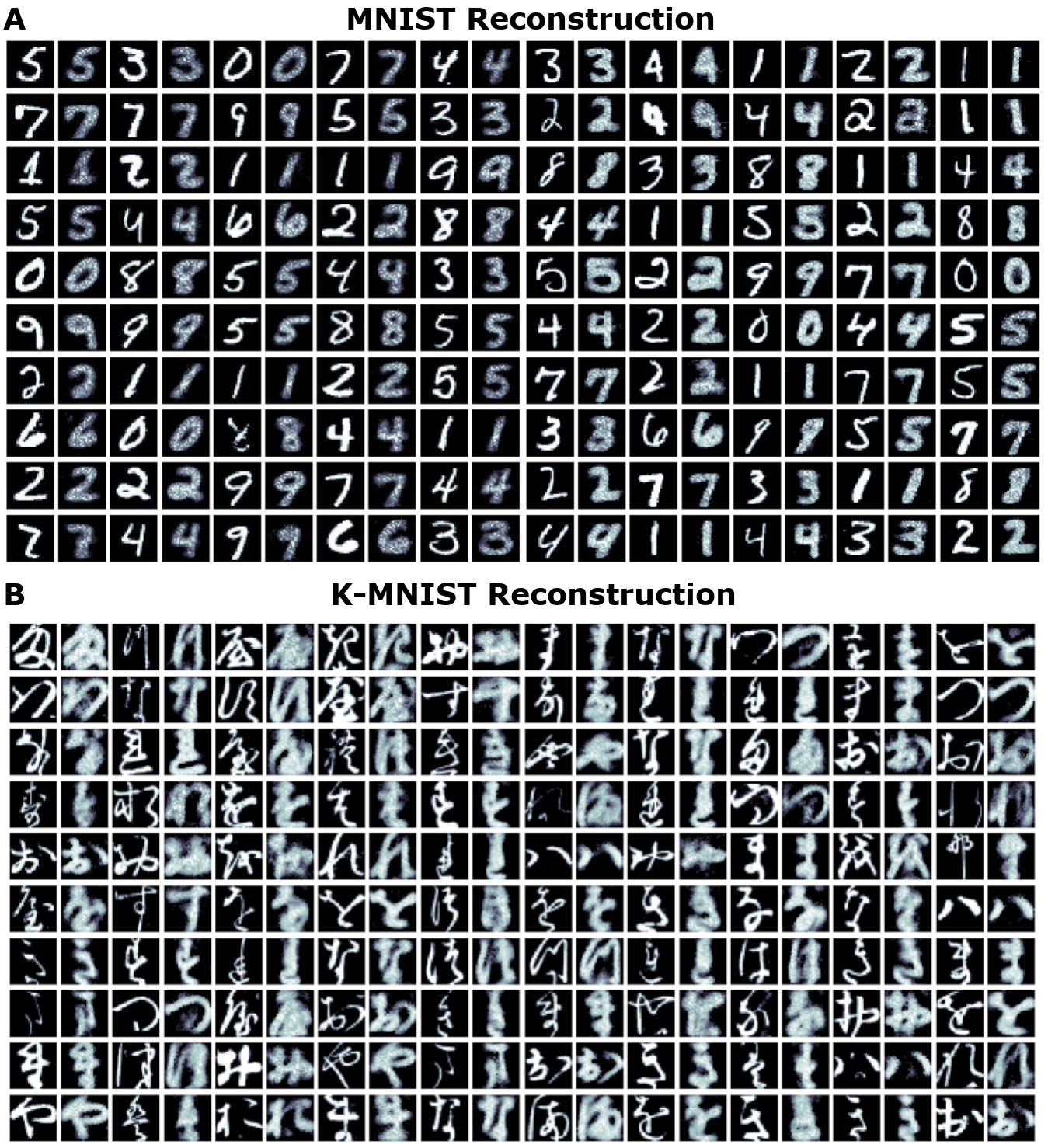}
        \caption{ \textbf{Reconstruction ability of a CSDP-trained SNN.} 
        Reconstructed samples produced by the SNN when learned with CSDP. {
        The top row (\textbf{A}) presents intercalated original images taken from the MNIST database (randomly sampled) with CSDP model reconstructed images; columns alternate between original and reconstruction patterns (starting with original images in the leftmost column). The bottom row (\textbf{B}) presents original images and reconstructed values for the K-MNIST database (formatted in the same way as sub-figure \textbf{A}).} }
        \label{fig:recon_viz}
        \vspace{-0.5cm}
\end{figure}

\begin{figure}[!t]
     \centering
    \includegraphics[width=0.875\textwidth]{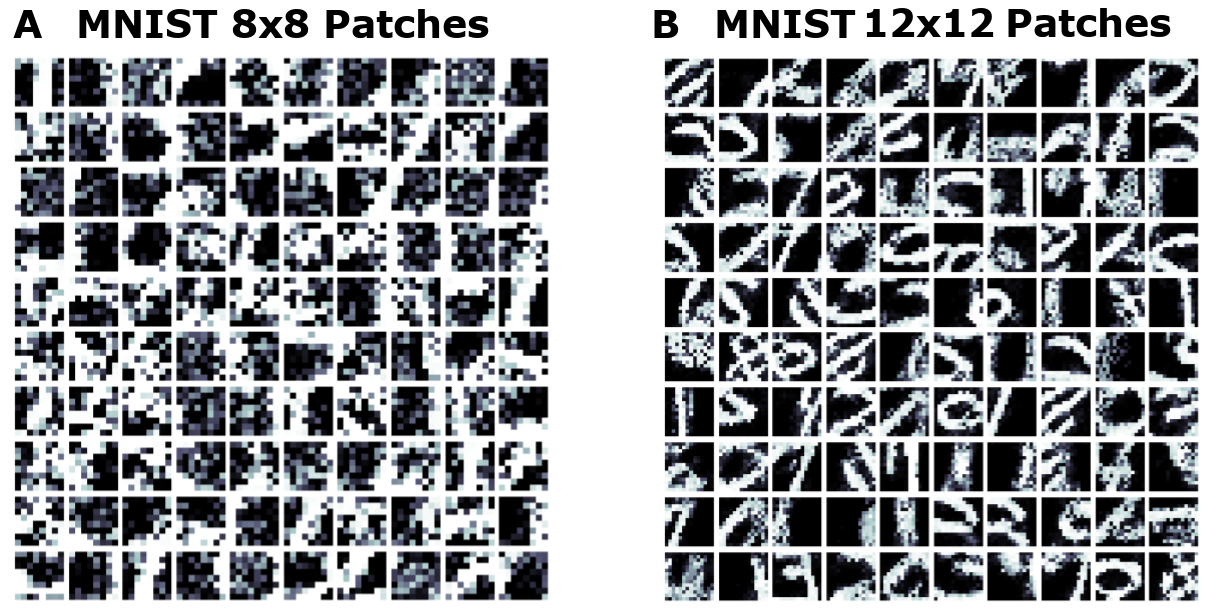}
        \caption{ \textbf{Bottom-layer receptive fields acquired by patch-level CSDP-trained SNNs.} 
        The receptive fields of $100$ randomly chosen LIF neuronal units, in layer $\ell = 1$, of unsupervised CSDP models, trained on \textcolor{black}{trained on} MNIST sample patches. Shown are the receptive fields for: ({\textbf{A}}) $8 \times 8$ patches, and ({\textbf{B}}) $12 \times 12$ patches.}
        \label{fig:patch_viz}
\end{figure}

\section*{Discussion} 


In this study, the proposed process of contrastive-signal-dependent plasticity (CSDP) offers a powerful yet simple neurobiologically-motivated scheme for local self-supervised learning in spiking neural systems. A central finding of our experiments is that iterative (contrastive) learning in a recurrent spiking neuronal circuit with fully parallel neural layer dynamics is possible. This critically resolves the core locality and locking problems, which are also sources of biological implausibility, that plague inference and credit assignment in many models of spiking neural networks, which often {rely on} forward transmission/inference neuronal circuitry similar to the architectures of modern-day deep neural networks. An important implication of this result is that our framework could be used to guide the effective design of parallel neuromorphic hardware arrangements with asynchronous updating of spiking neuronal dynamics.  
Furthermore, our results demonstrate that CSDP adaptation successfully learns temporal representations that facilitate effective classification as well as reconstruction of visual sensory inputs -- both unsupervised and supervised variants of our circuitry result in spiking neuronal activities that facilitate generative and discriminative capabilities. 
{Notably, CSDP is a viable candidate for instantiation in memristive platforms \cite{itoh2008memristor,thomas2013memristor}, given that it only requires pre-synaptic spike and post-synaptic spike/trace information (for how a CSDP circuit might be instantiated on a crossbar, see the supplementary text). Given that CSDP can be decomposed into a modulatory signal and a pre-synaptic STDP term (see supplementary text), one could take advantage of the long-standing work in emulating STDP and integrate-and-fire functionality through conventional CMOS-technology, possibly using $1$-bit analogue-digital converters, e.g., oversampled Delta-Sigma modulator units, in crafting CSDP-adapted NPUs.}

From a computational neuroscience point-of-view, our results suggest that a self-supervised contrastive signal could play an important role in local plasticity. This is noteworthy given that it is not guaranteed that feedback pathways or dopamine reward signals are always available to every region of neurons in the brain at every point in time{. CSDP demonstrates} how synaptic change might be carried out without the presence of feedback synaptic pathways to produce modulatory signals/perturbations or to pass along error mismatch messages (as in spiking predictive coding schemes \cite{rao2004hierarchical,ororbia2019spiking}). This could prove useful in modeling adaptivity in neuronal circuitry without the presence of globally broadcasted neuromodulation signals based on error or reward values, as in multi-factor Hebbian plasticity-based modeling frameworks \cite{izhikevich2007solving,kusmierz2017learning}. {Furthermore, CSDP strongly complements prior work on neuromimetic models based on contrastive Hebbian learning \cite{zheng2022correcting}, offering a decoupled and non-conditioned form of positive (plus) and negative (minus) phases of plasticity, facilitating the creation of spike-level contrastive credit assignment that could generalize schemes that focus on emulating biological wake phase learning and NREM sleep-induced replay \cite{singh2022sleep}.}
CSDP fundamentally operates on information locally available to synapses, in an architecture of fully in-parallel operating neuronal layers without requiring differentiable neural activities (which desirably side-steps the need for surrogate functions, which are required by schemes such as backpropagation applied to spiking networks \cite{lee2016training}), providing a viable candidate inference and learning scheme for thermodynamically efficient processing-in-memory and the instantiation of neuromorphic mortal computers.


\subsection*{On Limitations}
Although our online contrastive forward-only credit assignment process for training recurrent, layer-wise parallel SNNs is promising, there are several limitations to consider. {First, there are multiple hyperparameter constants to set/configure, and while CSDP models were found to be reasonably robust to most of these in this work, some consideration of the goodness threshold, the learning rate, and the synaptic decay factor should be given in practice (see the supplementary material for further discussion). }
Next, although the spiking model that we designed operates with bounded synaptic values (between $-1,1$), it does not operate with strictly positive values, which is an important hallmark of neurobiology -- we cannot have negative synaptic efficacies. This is an unfortunate limitation that is made worse by the fact that the signs of the synaptic values could change throughout the course of simulated learning. {Ultimately, this is in violation of Dale's Law \cite{sprekeler2017functional} as synapses emanating from the same neuron should all be excitatory or all inhibitory and not a mix.} 
Nevertheless, we remark that this issue could be potentially rectified by constraining all synapses to $[0,1]$ and then introducing an additional set of spiking inhibitory (or inter-inhibitory) neurons coupled to each layer which would provide the inhibitory/depression signals generally offered by negative synaptic strengths. The proportion of inhibitory neurons to excitatory ones could then {further} be desirably configured to adhere to {the ratios of excitatory to inhibitory neurons as reported in work in neurobiology and/or further to maintain a form of excitatory-inhibitory balance}. 
Future work will explore this possible reformulation, as the countering pressure offered by negative synapses is important for a goodness-based contrastive objective such as CSDP to work optimally -- ideally, inhibitory neurons should provide the require\textcolor{black}{d} type of counter-pressure.
\textcolor{black}{Finally, we remark that, while CSDP performed well on the datasets that we investigated, the performance gain(s) observed when going from the smaller (`Sm', with a bit over $7$ million plastic synapses) to larger models (`Lg', with a bit over $40$ million synapses) was rather modest, particularly on MNIST with less than a percentage point increase (in contrast, on K-MNIST, there was more than a $3$\% increase when increasing model size/complexity). This possibly indicates that the additional capacity afforded by the extra synaptic parameters is not being as effectively utilized by the learning process as it could be.  In the supplement, further evidence that supports this possibility is provided by an architecture size experiment that we conducted -- a law of diminishing returns effect was observed as the model was scaled to larger sizes/capacities (saturating to about $97.67$\% test accuracy on MNIST for the biggest model). 
Potential future improvements to CSDP-based model generalization, particularly on image-based sensory inputs, would likely come from the design and integration of useful inductive biases, e.g., convolution or locally-connected synaptic structures that approximate convolution, rather than just allocating a greater model capacity.} 

Beyond its use of synapses without a sign constraint, our current recurrent SNN does not include refractory periods, only implementing an instantaneous form of depolarization for all NPUs. However, as mentioned earlier in this article, this could be easily corrected by modifying our spike output functions 
to include an absolute refractory period, as in \cite{ororbia2019spiking}, as well as a relative refractory period. Furthermore, the adaptive thresholds that we presented in the main paper 
are calculated as a function of the total number of spikes across a layer $\ell$, at time $t$, whereas it would be more bio-physically realistic to have each leaky integrator unit within a layer adapt its own specific scalar threshold. Note that a per-neuron adaptive threshold could be modeled by another ordinary differential equation. 

With respect to the contrastive-signal-dependent plasticity  process itself, while we were able to desirably craft a simple update rule that operated at the spike-level, there is still the drawback that the rule, much like its rated-coded sources of inspiration \cite{hinton2022forward,ororbia2022predictive}, requires positive and negative data samples to compute meaningful synaptic adjustments. 
In this study, we made use of two simple approaches to synthesize negative samples: 
\textbf{1)} in the case of the supervised model variant, we exploited the fact that labels were readily available to serve as top-down {target class} signals which made synthesizing negative samples easy -- all we needed to do was simply select one of the incorrect class labels to create a negative {target class}, or 
\textbf{2)} {in the case of the unsupervised model variant, we generated} out-of-distribution patterns by interpolating between distinct pattern pairs within a mini-batch in tandem with randomly applied (image) rotation. 
Although our approaches for synthesizing negative data samples are simple and performed on-the-fly, investigating alternative, more sophisticated schemes would prove fruitful. One promising way to do this could be through the use of the predictive/generative synapses of the CSDP spike model we introduced earlier{. In other words,} the system could dynamically produce data confabulations that would serve as on-the-fly negative patterns. However, the greatest difficulty in synthesizing negative samples via the generative circuitry of our spiking model would be in crafting a process by which ancestral sampling could be efficiently conducted. We remark that crafting an ancestral sampling process in the context of spike-trains is not nearly as clear as it is in the realm of rate-coded models, such as in \cite{ororbia2022predictive}. 
Likely, one would need to develop a sampling scheme that is temporal in nature, which might be challenging to formulate in an online, efficient manner.

For the supervised variant of our model, an alternative to synthesizing negative samples would be to design another neural circuit that produces {(both positive and negative variations of) a} context vector {in place of} $\mathbf{y}$ instead of using a provided label -- a direction that is also more biologically realistic. Another drawback of using positive and negative samples in the kind of contrastive learning that we do in this work is that both types of data are used simultaneously. In \cite{hinton2022forward,ororbia2022predictive}, it has been discussed that it will be important to examine alternative schedules for when negative samples are presented to and used by a neural system to adjust its synapses. 
Another notable challenge facing CSDP, as well as other forms of biological credit assignment, is that of scaling. While this was not explored in this study, CSDP would be amenable to the use of operations such as convolution/deconvolution as well as operate in the context of architectures of depth greater than the four-layer models investigated in this work. Given that CSDP updates only require incoming (pre-synaptic spikes) information and outgoing values (e.g., post-synaptic information and a cross-layer modulatory signal), utilizing operations useful for processing more complex sensory inputs, e.g., natural images, should be possible. One key difficulty in scaling CSDP, particularly the unsupervised variant, will be in the design of useful negative samples, which will likely be task or data type dependent (this work took advantage of the fact that the inputs were pixel images and were thus amenable to image/pixel transformations; other forms of data would require other means of generating negative data). 
It would be fruitful for future work to investigate CSDP's efficacy in adapting more complex spiking neuronal architectures (as well as other types of spiking neuronal dynamics, e.g., as in Izhikevich cells) \textcolor{black}{to} a wider variety of problems contexts.

Finally, although CSDP results in synaptic updates that \textcolor{black}{are} local in terms of neural circuit structure (updates for each layer's synaptic bundles happen in parallel), the contrastive modulator $\delta^\ell(t)$ central to CSDP is spatially non-local in \textcolor{black}{and} of itself. In other words, it requires activity trace information taken across a layer/group of neurons since the goodness probability is computed via summation across trace values. Biologically, a form of this modulatory signal could come from astrocytic cellular support \cite{santello2009synaptic,han2013forebrain}. This would mean that each value within $\delta^\ell(t)$, i.e., $\delta^\ell_i(t)$ for any neuron $i$ in layer $\ell$, is produced via a ``helper cell'' that can quickly aggregate readily available information across a group of spatially nearby spiking neuronal cells, \textcolor{black}{embodying the production of transporters for various neurotransmitters} \cite{santello2009synaptic} 
or serving as the impetus \textcolor{black}{for} an increase in long-term potentiation  \cite{han2013forebrain}. In the supplement, we provide one possible neuromorphic implementation of this supporting astrocytic computation.

\section*{Materials and Methods}

\subsection*{Datasets Utilized} 
The data utilized in for the simulations conducted in this study came from the MNIST and Kuzushiji-MNIST (K-MNIST) databases. 
The MNIST dataset \cite{lecun1998mnist} 
specifically contains images of handwritten digits across $10$ different categories. Kuzushiji-MNIST (K-MNIST) is a challenging drop-in replacement for MNIST, containing images depicting hand-drawn Japanese Kanji characters \cite{clanuwat2018deep}{. Each} class in this database corresponds to the character's modern hiragana counterpart, with $10$ total classes. For both datasets, image patterns were normalized to the range of $[0,1]$ by dividing pixel values by $255$. The resulting pixel ``probabilities'' were then used to create sensory input spike trains by treating the normalized vector $\mathbf{x}/255$ as the parameters for a multivariate Bernoulli distribution. The resulting distribution was sampled at each time step $t$ over the stimulus window of length $T$. Note that we did not preprocess the image data any further unlike previous efforts related to spiking networks. However, we remark that it might be possible to obtain better performance by whitening image patterns (particularly in the case of natural images) or applying a transformation that mimics the result of neural encoding populations based on Gaussian receptive fields. 
{From both MNIST and K-MNIST, which each contained $60,000$ training samples (and a test-set of $10,000$ samples), a validation/development subset of $10,000$ patterns ($1,000$ from each class) was created by randomly sampling without replacement from each database’s training set (the development subset was used to manually tune/select hyperparameter values). These subsets were used to \textcolor{black}{aid in} selecting/verifying constants/hyperparameters.
}

\subsection*{Spiking Model Specification Details}
All of the synapses of the full model, composed of $L$ layers of neural processing elements (NPUs), can be represented as a set of matrices contained in the synaptic parameter construct $\Theta = \{\mathbf{W}^1,\mathbf{V}^1,\mathbf{M}^1,\mathbf{B}^1,...,\mathbf{W}^\ell,\mathbf{V}^\ell,\mathbf{M}^\ell,\mathbf{B}^\ell,...\mathbf{W}^L,\mathbf{M}^L,\mathbf{B}^L \}$ (the top layer $L$ does not contain any top-down recurrent synapses as there would be no layer above $L$). 
Matrix $\mathbf{W}^\ell \in [-1,1]^{J_\ell \times J_{\ell-1}}$ contains the local bottom-up synaptic connections, $\mathbf{V}^\ell \in [-1,1]^{J_\ell \times J_{\ell+1}}$ contains the local top-down recurrent synapses, $\mathbf{M}^\ell \in [0,1]^{J_\ell \times J_\ell}$ holds the lateral inhibition synapses, and the $\mathbf{B}^\ell \in [-1,1]^{J_\ell \times C}$ stores the optional {target class}-mediating connections. The values of all synaptic strengths in our model are constrained to remain in the range $[-1,1]$ except for the lateral ones, which are constrained to non-negative values $[0,1]$. 

In accordance with our CSDP scheme, the Hebbian-like adjustment rules for all possible synaptic connections that project to layer $\ell$ are specifically:
\begin{equation}
    \delta^\ell_i(t) =  \partial \mathcal{C}(\mathbf{z}^\ell(t), y_{type}) \Big/ \partial z^\ell_i(t)
\end{equation}
\begin{equation}
    \Delta W^\ell_{ij} = \Big( R_E \delta^\ell_i(t)  s^{\ell-1}_j(t-\Delta t)  \Big) + \lambda_d \Big( s^\ell_i(t) \big(1 - s^{\ell-1}_j(t - \Delta t)\big) \Big)
\end{equation}
\begin{equation}
    \Delta V^\ell_{ij} = \Big( R_E \delta^\ell_i(t) s^{\ell+1}_j(t-\Delta t)  \Big) + \lambda_d \big( s^\ell_i(t) \big(1 - s^{\ell+1}_j(t - \Delta t)\big) \big)
\end{equation}
\begin{equation}
    \Delta M^\ell_{ij} = \Big( R_I \delta^\ell_i(t) s^{\ell}_j(t-\Delta t)  \Big) + \lambda_d \Big( s^\ell_i(t) \big(1 - s^{\ell}_j(t-\Delta t)\big) \Big)
\end{equation}
\begin{equation}
    \Delta B^\ell_{ij} = \big( R_E \delta^\ell_i(t)  s_{y,j}(t-\Delta t)  + \lambda_d \Big( s^\ell_i(t)  \big(1 - s_{y,j}(t-\Delta t)\big) \Big).
\end{equation}
In this study, we set the decay factor for the second term of our local adjustment rule to be $\lambda_d = 0.00005$ (typically smaller values were found to result in the best performance). 
In the supplementary material, we provide a pseudocode that fully depicts the mechanics of the inference and learning underlying a CSDP-trained spiking circuit.

\begin{figure}[!t]
    \begin{center}
    \includegraphics[width=0.9\textwidth]{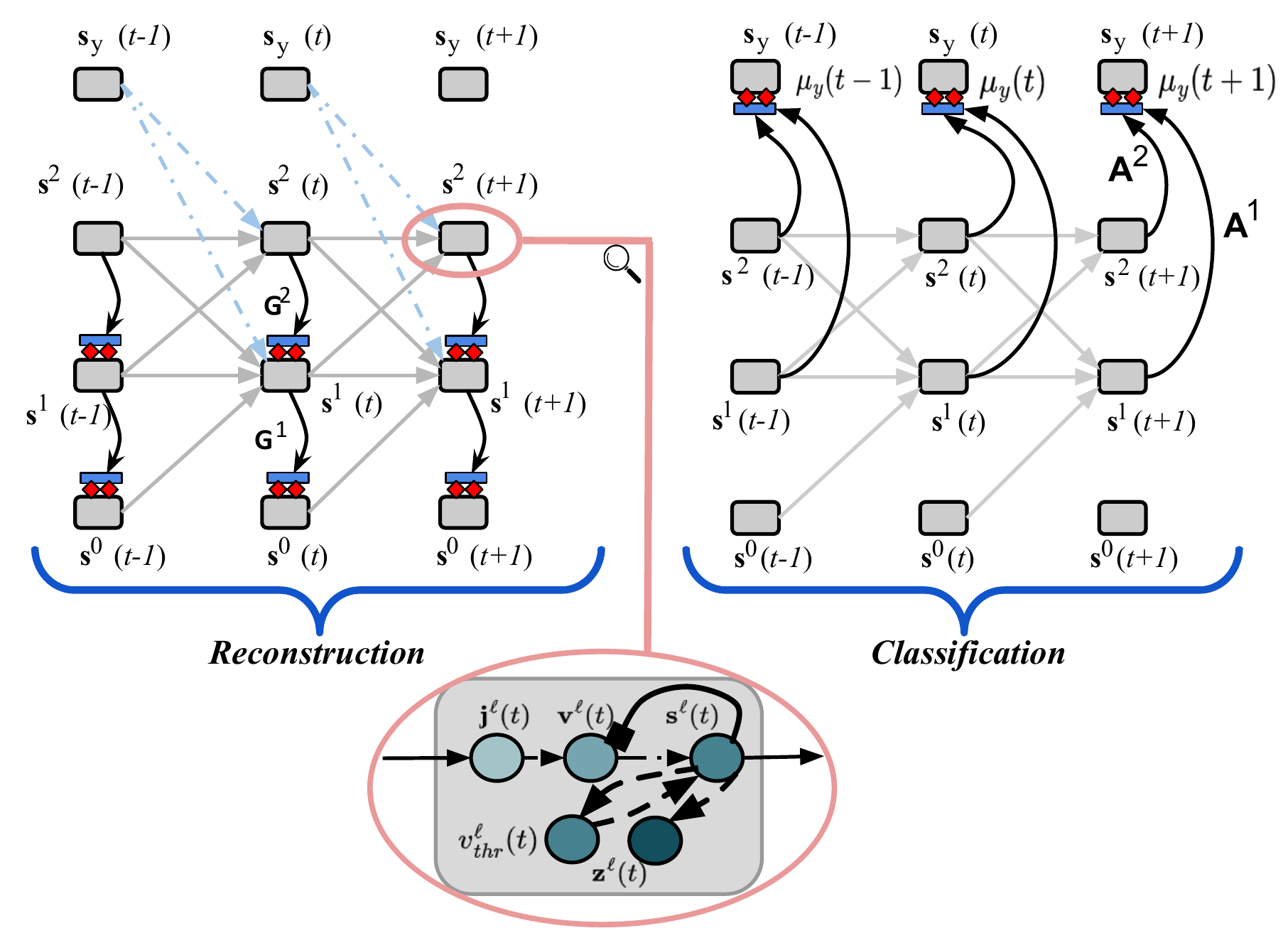}
    \end{center}
    \vspace{-0.2cm} 
        \caption{ \textbf{Reconstruction and classification processes of a CSDP-trained SNN.} 
        {An overview of the task-specific extensions made to the} CSDP-adapted recurrent spiking network, unfolded over three time steps, which processes a sensory spike train $\{\mathbf{s}^0(0),...,\mathbf{s}^0(t),...,\mathbf{s}^0(T)\})$ with possibly an available corresponding {target class} spike train $\{\mathbf{s}_y(0),...,\mathbf{s}_y(t),...,\mathbf{s}_y(T)\})$. The zoomed-in inset depicts, internally, that each activation layer is made up of at least four components, i.e., an electrical current model $\mathbf{j}^\ell(t)$, a voltage model $\mathbf{v}^\ell(t)$, a spike response function $\mathbf{s}^\ell(t)$, and an adaptive threshold $v^\ell_{thr}(t)$. {The leftmost unfolded model (reconstruction) introduces additional synaptic bundles $\mathbf{G}^1$ and $\mathbf{G}^2$ that allow a layer $\ell$ to make local predictions of the neuronal activities of layer $\ell-1$. The rightmost unfolded model (classification) introduces extra synaptic connections $\mathbf{A}^1$ and $\mathbf{A}^2$ that allow the layers to make a prediction $\mathbf{\mu}_y(t)$ of the target class spikes $\mathbf{s}_y(t)$ at time $t$. } Note that the light dot-dashed arrows, representing the class-modulating synaptic connections, are not used in the case of the unsupervised model's reconstruction process (since class context is not available for this case).
        }
        \label{fig:csdp_snn_arch}
        \vspace{-0.5cm}
\end{figure} 

\subsubsection*{A CSDP-SNN as a Spiking Generative Model}
To incorporate  reconstruction/generative potential into our SNN model, we generalize the local generation scheme of the predictive forward-forward (PFF) scheme \cite{ororbia2022predictive} and introduce one more set of matrices which contain the generative synapses -- this means that we extend our model parameters to $\Theta \bigcup \Theta_g$ where $\Theta_{g} = \{\mathbf{G}^\ell\}^L_{\ell=1}$. Desirably, each synaptic matrix $\mathbf{G}^\ell \in [-1,1]^{J_{\ell} \times J_{\ell+1}}$ is directly wired to each layer $\ell$, forming a local prediction of activity $\mathbf{s}^\ell(t)$ {as follows}:
\begin{equation}
    \mathbf{v}^\ell_\mu(t + \Delta t) = \mathbf{v}^\ell_\mu(t) + (\Delta t/\tau_m) \big( -\mathbf{v}^\ell_\mu(t) + R_E (\mathbf{G}^{\ell+1} \cdot \mathbf{s}^{\ell+1}(t)) \big) \label{eqn:recon_voltage}
\end{equation}
\begin{equation}
    \mathbf{s}^\ell_\mu(t) = \mathbf{v}^\ell_\mu(t + \Delta t) > v^\ell_{thr,\mu}, \quad \mathbf{v}^\ell_\mu(t) = \mathbf{v}^\ell_\mu(t+ \Delta t) \odot (1 - \mathbf{s}^\ell_\mu(t) ) \label{eqn:recon_spike_model}
\end{equation}
\begin{equation}
    \mathbf{e}^\ell(t) = \mathbf{s}^\ell_\mu(t) - \mathbf{s}^\ell(t) \label{eqn:mismatch}
\end{equation}
where {$\odot$ denotes elementwise multiplication and $\cdot$ denotes matrix-vector multiplication}.  
We observe that error mismatch units $\mathbf{e}^\ell(t)$ have been introduced that specialize in tracking the disparity between prediction spikes $\mathbf{s}^\ell_\mu(t)$ and a current representation spikes $\mathbf{s}^\ell(t)$. To update each generative matrix $\mathbf{G}^\ell$, we may then use the simple Hebbian update rule:
\begin{equation}
    \Delta \mathbf{G}^\ell = \big( R_E \mathbf{e}^{\ell-1}(t) \big) \cdot (\mathbf{s}^{\ell}(t))^T 
\end{equation}
\textcolor{black}{where $\big(\mathbf{s}^\ell(t)\big)^T$ indicates that the transpose operator is applied to $\mathbf{s}^\ell(t)$.} 
When making local predictions and calculating mismatches as above, the full spiking neural system's {optimization objective},  
at time $t$, can be written down as:
\begin{equation}
    \mathcal{F}(t,\Theta) = \sum^{L}_{\ell=0}
    \begin{dcases}
        \mathcal{C}(\mathbf{z}^\ell(t), y_{type}) & \text{if } \ell = L \\
        (1/2)||\mathbf{s}^\ell_\mu(t) - \mathbf{s}^\ell(t)||^2_2 & \text{if } \ell = 0 \\
        \mathcal{C}(\mathbf{z}^\ell(t), y_{type}) + (1/2)||\mathbf{s}^\ell_\mu(t) - \mathbf{s}^\ell(t)||^2_2 & \text{otherwise.}
    \end{dcases} 
    \label{eqn:pff_local_cost}
\end{equation}
Notice that each layer of the system is maximizing its goodness while further learning a (temporal) mapping between its own spike train to the one of the layer below it (i.e., a mapping that minimizes a measurement of local predictive spike error). 
The mismatch units we model explicitly in Equation 20 
directly {follow} from taking the partial derivative of the local objective in Equation 22 
with respect to $\mathbf{s}^\ell_\mu(t)$. Finally, across a full stimulus window of length $T$, the global objective that our spiking system would be optimizing is the following sequence loss:
\begin{equation}
    \mathcal{F}(\Theta) = \sum^T_{t=1} \mathcal{F}(t,\Theta) \mbox{.} \label{eqn:seq_fe}
\end{equation}
In effect, the spiking neural system that we simulate in this work attempts to incrementally optimize the sequence loss in Equation 23 
when processing a data sample $(\mathbf{x},\mathbf{y})$ (or $\mathbf{x}$ in the case of unsupervised learning setups), conducting a form of online adaptation by computing synaptic adjustments given the current state of $\Theta$ at time $t$. In Fig. 7 A 
we depict how the local (spike) predictions are made in our recurrent spiking architecture. 

\subsubsection*{Pattern Reconstruction Details}
For a given CSDP-trained SNN, a single reconstructed pattern $\mathbf{\widehat{x}}$ was created by computing an average across (clipped) activation traces of the bottom-most predictor's output spikes, collected during the $T$-length stimulus window as follows:
\begin{equation}
    \mathbf{\widehat{x}} = (1/T) \sum^T_{t=1} \mathbf{z}^0_\mu(t), \; \mbox{where } \mathbf{z}^0_\mu(t) = \big( \mathbf{z}^0_\mu + (\Delta t/\tau_{tr}) (-\mathbf{z}^0_\mu(t)) \big) \odot (1 - \mathbf{s}^0_\mu(t)) + \mathbf{s}^0_\mu(t)
\end{equation}
where $\mathbf{z}^0_\mu(t)$ is the activation trace of the output prediction spike vector $\mathbf{s}^0_\mu(t)$, produced by the local prediction of layer $\ell=0$ from layer $\ell=1$ (see Equation 18
), at simulation step $t$. 
{Note that for the reconstruction analysis, the class $\mathbf{y}$ was not clamped in our model for reconstruction to be evaluated. This means that the model could only employ {class-specific} information it had previously stored in its synapses to determine how to best reconstruct test samples.}

\subsubsection*{Spike-Driven Classification and its Fast Approximation}
To perform classification with the recurrent spiking system model developed over the previous few sections, one must run a scheme similar to \cite{hinton2022forward,ororbia2022predictive} where one iterates over all possible class values, i.e., setting the {target} vector $\mathbf{y}$ equal to the one-hot encoding of each class $c \in \{1,2,...,C\}$. Specifically, for each possible class index $c$, we create a candidate target/class vector $\mathbf{y} = \mathbf{1}_c${. $\mathbf{1}_c$ is the indicator function, returning one when the index $i$ (of the label space) corresponds to the correct class index $c \in \{1,2,...,C\}$ and zero otherwise. We then present $\mathbf{y}$} {and the sensory input} $\mathbf{x}$ to the system over a stimulus window of $T$ steps \textcolor{black}{and} record the goodness values summed across layers, averaged over time, i.e., $\mathcal{G}_{y=c} = (1/T) \sum^T_{t=1} \sum^{L}_{\ell=1} \sum^{J_\ell}_{j=1} ( \mathbf{z}^\ell_j(t) )^2$. This time-averaged goodness value is computed for all class indices $y = 1, 2,...,C$, resulting in an array of $C$ scores, i.e, $G = \{ \mathcal{G}_{y=1}, \mathcal{G}_{y=2}, ..., \mathcal{G}_{y=C} \}$. The argmax operation is applied to this array $G$ to obtain the index of the class with the highest average (aggregate) goodness value, i.e., $c_{pred} = \argmax_{c \in C} G$. 

While the above per-class {classification} process would work fine for reasonable values of $C$, it would not scale well to a high number of classes, i.e., very large values of $C$, given that the above process requires simulating the spiking network over $C$ stimulus windows in order to compute a predicted label index that corresponds to maximum goodness. {To circumvent this expensive form of classification}, we constructed a simple modification which circumvents the need to conduct this classification process entirely by integrating a simple spiking classification sub-circuit to the recurrent spiking system. This classifier, which takes in as input the spike output produced at the top level of the recurrent system, i.e., $\mathbf{s}^L(t)$, is jointly learned with the rest of the model parameters{. This} merely entails extending $\Theta$ one more time so as to include classification-specific synapses, i.e., $\Theta \bigcup \Theta_c$ where $\Theta_c = \{\mathbf{A}^\ell\}^L_{\ell=1}$. Concretely, the classifier module operates according to the following dynamics:
\begin{equation}
    \mathbf{v}_y(t + \Delta t) = \mathbf{v}_y(t) + (\Delta t/\tau_m) \Big( -\mathbf{v}_y(t) + R_E \Big( \sum^L_{\ell=1} \mathbf{A}^\ell \cdot \mathbf{s}^\ell(t) \Big) \Big) \label{eqn:context_voltage}
\end{equation}
\begin{equation}
    \mathbf{\mu}_y(t) = \mathbf{v}_y(t + \Delta t) > v^y_{thr}, \; \mathbf{v}_y(t) = \mathbf{v}_y(t+ \Delta t) \odot (1 - \mathbf{\mu}_y(t) ) \label{eqn:context_spike_model}
\end{equation}
\begin{equation}
    v^y_{thr} = v^y_{thr} + \lambda_v \Big( (\sum^{C}_{j=1} \mathbf{\mu}_{y,j}(t) - 1 \Big)  \label{eqn:context_threshold}
\end{equation}
where we notice, in Equation 25, 
that the {class} prediction voltage $\mathbf{v}_y(t)$ is the result of aggregating across spike vector messages transmitted from layers $\ell=1$ to $\ell=L$.   
Each synaptic matrix $\mathbf{A}^\ell$ that makes up the spiking classifier is adjusted according to the following Hebbian rule:
\begin{equation}
    \Delta \mathbf{A}^\ell = R_E \big( \mathbf{\mu}_y(t) - \mathbf{s}_y(t) \big) \cdot (\mathbf{s}^\ell(t))^T \mbox{.}
\end{equation} 
Note that, in the case of unsupervised CSDP, $\mathbf{s}_y(t)$ is not provided as class contextual input (the class-modulating synapses are zeroed out/removed)\textcolor{black}{, even if the model is equipped with the above skip-layer classification synapses (i.e., the class spike error signal only affects the plasticity of the classification synapses $\{\mathbf{A}^\ell\}^L_{\ell=1}$)}. 
If these additional synaptic classification parameters are learned, then, at test time, when synaptic adjustment would be disabled, one can simply provide the spiking system with the sensory input $\mathbf{x}$ with no target class $\mathbf{y}$ and obtain a estimated {class} output $\mathbf{\bar{y}}$. This $\mathbf{\bar{y}}$ is a function of the system's predictions across time, i.e., it is the approximate output distribution computed as: $p(\mathbf{y}|\mathbf{x};\Theta) \approx \mathbf{\bar{y}} = \exp (  \sum^T_{t=1} \mathbf{\mu}_y(t) ) / \sum^C_{c=1} \exp (  \sum^T_{t=1} \mathbf{\mu}_y(t) )_c $. Note that the conditional log likelihood $\log p(\mathbf{y}|\mathbf{x};\Theta)$ of the spiking model can now be measured using this approximate label probability distribution. Although learning this spiking classifier entails a memory cost of up to $L$ additional synaptic matrices, the test-time inference of the spiking system will be substantially faster, thus making it more suitable for on-chip computation. 
In Fig. 7 
B, 
our recurrent spiking neural architecture is shown making predictions of the target class's spike train across three time steps.

\section*{Acknowledgments}
We would like to thank Alexander Ororbia (Sr.) for useful feedback and revisions on the draft of the manuscript. \\
\textbf{Funding:} We acknowledge the support of the Cisco Research Gift Award \#26224. \\
\textbf{Authors contributions:} \textit{Conceptualization, Methodology, Investigation, Writing:} AO (AO conceptualized the architecture and plasticity scheme, developed the mathematical models, wrote the simulation framework, performed the experiments, and wrote the paper.) \\
\textbf{Competing interests:} The authors declare that they have no competing interests. \\
\textbf{Data and materials availability:} All datasets used in this work are publicly available. All data needed to evaluate the conclusions in the paper are present in the paper and/or the Supplementary Materials. 
{The code for performing the simulations of this work was written in the Python programming language (Python \textcolor{black}{$3.10.6$}), further using the JAX numerical computing package (jax/jaxlib \textcolor{black}{$0.4.28$}) and the ngc-learn simulation library (ngclearn \textcolor{black}{$1.2$-beta$3$} and ngcsimlib \textcolor{black}{$0.3$-beta4}). 
} 
Code to run the neuronal simulations is available at the GitHub repository:\\
\url{https://github.com/ago109/contrastive-signal-dependent-plasticity} (Version \textcolor{black}{$0.2.0$}, Zenodo Identifier: \textcolor{black}{$13345741$}).

\bibliography{refs}

\begin{thebibliography}{10}

\bibitem{beyeler2013categorization}
{\sc Beyeler, M., Dutt, N.~D., and Krichmar, J.~L.}
\newblock Categorization and decision-making in a neurobiologically plausible
  spiking network using a stdp-like learning rule.
\newblock {\em Neural networks 48\/} (2013), pp. 109--124.

\bibitem{bi1998synaptic}
{\sc Bi, G.-q., and Poo, M.-m.}
\newblock Synaptic modifications in cultured hippocampal neurons: dependence on
  spike timing, synaptic strength, and postsynaptic cell type.
\newblock {\em Journal of neuroscience (J. Neurosci.) 18\/} (1998), pp.
  10464--10472.

\bibitem{brevini2020black}
{\sc Brevini, B.}
\newblock Black boxes, not green: Mythologizing artificial intelligence and
  omitting the environment.
\newblock {\em Big Data \& Society (BD\&S) 7\/} (2020), 2053--9517.

\bibitem{clanuwat2018deep}
{\sc Clanuwat, T., Bober-Irizar, M., Kitamoto, A., Lamb, A., Yamamoto, K., and
  Ha, D.}
\newblock Deep learning for classical japanese literature.
\newblock {\em arXiv:1812.01718\/} (2018).

\bibitem{clark2015surfing}
{\sc Clark, A.}
\newblock {\em Surfing uncertainty: Prediction, action, and the embodied mind}.
\newblock Oxford University Press, 2015.

\bibitem{cox2014neural}
{\sc Cox, D.~D., and Dean, T.}
\newblock Neural networks and neuroscience-inspired computer vision.
\newblock {\em Current biology (Curr. Biol.) 24}, 18 (2014), pp. 921--929.

\bibitem{diehl2015unsupervised}
{\sc Diehl, P.~U., and Cook, M.}
\newblock Unsupervised learning of digit recognition using
  spike-timing-dependent plasticity.
\newblock {\em Frontiers in computational neuroscience (Front. Comput.
  Neurosci.) 9\/} (2015), 99.

\bibitem{eliasmith2012large}
{\sc Eliasmith, C., Stewart, T.~C., Choo, X., Bekolay, T., DeWolf, T., Tang,
  Y., and Rasmussen, D.}
\newblock A large-scale model of the functioning brain.
\newblock {\em Science 338\/} (2012), pp. 1202--1205.

\bibitem{frenkel2019learning}
{\sc Frenkel, C., Lefebvre, M., and Bol, D.}
\newblock Learning without feedback: Direct random target projection as a
  feedback-alignment algorithm with layerwise feedforward training.
\newblock {\em arXiv:1909.01311\/} (2019).

\bibitem{grossberg1987competitive}
{\sc Grossberg, S.}
\newblock Competitive learning: From interactive activation to adaptive
  resonance.
\newblock {\em Cognitive science (Cogn. Sci.) 11\/} (1987), pp. 23--63.

\bibitem{gutig2016spiking}
{\sc G{\"u}tig, R.}
\newblock Spiking neurons can discover predictive features by aggregate-label
  learning.
\newblock {\em Science 351\/} (2016), aab4113.

\bibitem{han2013forebrain}
{\sc Han, X., Chen, M., Wang, F., Windrem, M., Wang, S., Shanz, S., Xu, Q.,
  Oberheim, N.~A., Bekar, L., Betstadt, S., Silva, A.~J., Takano, T., Goldman,
  S.~A., and Nedergaard, M.}
\newblock Forebrain engraftment by human glial progenitor cells enhances
  synaptic plasticity and learning in adult mice.
\newblock {\em Cell stem cell 12\/} (2013), pp. 342--353.

\bibitem{hao2020biologically}
{\sc Hao, Y., Huang, X., Dong, M., and Xu, B.}
\newblock A biologically plausible supervised learning method for spiking
  neural networks using the symmetric stdp rule.
\newblock {\em Neural networks 121\/} (2020), pp. 387--395.

\bibitem{hazan2018unsupervised}
{\sc Hazan, H., Saunders, D., Sanghavi, D.~T., Siegelmann, H., and Kozma, R.}
\newblock Unsupervised learning with self-organizing spiking neural networks.
\newblock {\em 2018 International Joint Conference on Neural Networks (IJCNN)
  2018\/} (2018), 1--6.

\bibitem{hinton2022forward}
{\sc Hinton, G.}
\newblock The forward-forward algorithm: Some preliminary investigations.
\newblock {\em arXiv:2212.13345\/} (2022).

\bibitem{hinton2006reducing}
{\sc Hinton, G.~E., and Salakhutdinov, R.~R.}
\newblock Reducing the dimensionality of data with neural networks.
\newblock {\em Science 313}, 5786 (2006), 504--507.

\bibitem{itoh2008memristor}
{\sc Itoh, M., and Chua, L.~O.}
\newblock Memristor oscillators.
\newblock {\em International journal of bifurcation and chaos (IJBC) 18\/}
  (2008), pp. 3183--3206.

\bibitem{izhikevich2007solving}
{\sc Izhikevich, E.~M.}
\newblock Solving the distal reward problem through linkage of stdp and
  dopamine signaling.
\newblock {\em Cerebral cortex 17}, 10 (2007), pp. 2443--2452.

\bibitem{jaderberg2017decoupled}
{\sc Jaderberg, M., Czarnecki, W.~M., Osindero, S., Vinyals, O., Graves, A.,
  Silver, D., and Kavukcuoglu, K.}
\newblock Decoupled neural interfaces using synthetic gradients.
\newblock {\em International conference on machine learning (ICML)\/} (2017),
  1627--1635.

\bibitem{kingma2014adam}
{\sc Kingma, D., and Ba, J.}
\newblock Adam: A method for stochastic optimization.
\newblock {\em arXiv:1412.6980\/} (2014).

\bibitem{kohan2023signal}
{\sc Kohan, A., Rietman, E.~A., and Siegelmann, H.~T.}
\newblock Signal propagation: The framework for learning and inference in a
  forward pass.
\newblock {\em IEEE Transactions on Neural Networks and Learning Systems (IEEE
  Trans. Neural Netw. Learn. Syst.) 2023\/} (2023), 1--12.

\bibitem{kohan2018error}
{\sc Kohan, A.~A., Rietman, E.~A., and Siegelmann, H.~T.}
\newblock Error forward-propagation: Reusing feedforward connections to
  propagate errors in deep learning.
\newblock {\em arXiv:1808.03357\/} (2018).

\bibitem{konkle2022self}
{\sc Konkle, T., and Alvarez, G.~A.}
\newblock A self-supervised domain-general learning framework for human ventral
  stream representation.
\newblock {\em Nature communications (Nat. Commun.) 13\/} (2022), 491.

\bibitem{kusmierz2017learning}
{\sc Ku{\'s}mierz, {\L}., Isomura, T., and Toyoizumi, T.}
\newblock Learning with three factors: modulating hebbian plasticity with
  errors.
\newblock {\em Current opinion in neurobiology (Curr. Opin. Neurobiol.) 46\/}
  (2017), pp. 170--177.

\bibitem{lapicque1907recherches}
{\sc Lapicque, L.}
\newblock Recherches quantitatives sur l’excitation electrique des nerfs.
\newblock {\em Journal of Physiol Pathol G{\'{e}}n{\'{e}}rale 9\/} (1907), pp.
  620--635.

\bibitem{lecun1998mnist}
{\sc LeCun, Y.}
\newblock The mnist database of handwritten digits.
\newblock {\em http://yann.lecun.com/exdb/mnist/\/} (1998).

\bibitem{lee2016training}
{\sc Lee, J.~H., Delbruck, T., and Pfeiffer, M.}
\newblock Training deep spiking neural networks using backpropagation.
\newblock {\em Frontiers in neuroscience (Front. Neurosci.) 10\/} (2016), 508.

\bibitem{luo2021architectures}
{\sc Luo, L.}
\newblock Architectures of neuronal circuits.
\newblock {\em Science 373\/} (2021), eabg7285.

\bibitem{maass1997networks}
{\sc Maass, W.}
\newblock Networks of spiking neurons: the third generation of neural network
  models.
\newblock {\em Neural networks 10}, 9 (1997), pp. 1659--1671.

\bibitem{merolla2011digital}
{\sc Merolla, P., Arthur, J., Akopyan, F., Imam, N., Manohar, R., and Modha,
  D.~S.}
\newblock A digital neurosynaptic core using embedded crossbar memory with 45pj
  per spike in 45nm.
\newblock {\em 2011 IEEE custom integrated circuits conference (CICC) 2011\/}
  (2011), pp. 1--4.

\bibitem{merolla2014million}
{\sc Merolla, P.~A., Arthur, J.~V., Alvarez-Icaza, R., Cassidy, A.~S., Sawada,
  J., Akopyan, F., Jackson, B.~L., Imam, N., Guo, C., Nakamura, Y., Brezzo, B.,
  Vo, I., Esser, S.~K., Appuswamy, R., Taba, B., Amir, A., Flickner, M.~D.,
  Rick, W.~P., Manohar, R., and Modha, D.~S.}
\newblock A million spiking-neuron integrated circuit with a scalable
  communication network and interface.
\newblock {\em Science 345\/} (2014), 668--673.

\bibitem{mostafa2018deep}
{\sc Mostafa, H., Ramesh, V., and Cauwenberghs, G.}
\newblock Deep supervised learning using local errors.
\newblock {\em Frontiers in neuroscience (Front. Neurosci.) 12\/} (2018), 608.

\bibitem{nayebi2023neural}
{\sc Nayebi, A., Rajalingham, R., Jazayeri, M., and Yang, G.~R.}
\newblock Neural foundations of mental simulation: Future prediction of latent
  representations on dynamic scenes.
\newblock {\em arXiv:2305.11772\/} (2023).

\bibitem{ndri2023}
{\sc N’Dri, A.~W., Barbier, T., Teulière, C., and Triesch, J.}
\newblock {Predictive Coding Light: learning compact visual codes by combining
  excitatory and inhibitory spike timing-dependent plasticity*}.
\newblock {\em 2023 IEEE/CVF Conference on Computer Vision and Pattern
  Recognition Workshops (CVPRW) 00\/} (2023), pp. 3997--4006.

\bibitem{oreilly2000computational}
{\sc O'Reilly, R.~C., and Munakata, Y.}
\newblock {\em Computational explorations in cognitive neuroscience:
  Understanding the mind by simulating the brain}.
\newblock MIT press, 2000.

\bibitem{ororbia2019spiking}
{\sc Ororbia, A.}
\newblock Spiking neural predictive coding for continually learning from data
  streams.
\newblock {\em Neurocomputing 544\/} (2023), 126292.

\bibitem{ororbia2023mortal}
{\sc Ororbia, A., and Friston, K.}
\newblock Mortal computation: A foundation for biomimetic intelligence.
\newblock {\em arXiv preprint arXiv:2311.09589\/} (2023).

\bibitem{ororbia2022predictive}
{\sc Ororbia, A., and Mali, A.}
\newblock The predictive forward-forward algorithm.
\newblock {\em arXiv:2301.01452\/} (2022).

\bibitem{patterson2021carbon}
{\sc Patterson, D., Gonzalez, J., Le, Q., Liang, C., Munguia, L.-M., Rothchild,
  D., So, D., Texier, M., and Dean, J.}
\newblock Carbon emissions and large neural network training.
\newblock {\em arXiv:2104.10350\/} (2021).

\bibitem{pei2019towards}
{\sc Pei, J., Deng, L., Song, S., Zhao, M., Zhang, Y., Wu, S., Wang, G., Zou,
  Z., Wu, Z., He, W., Chen, F., Deng, N., Wu, S., Wang, Y., Wu, Y., Yang, Z.,
  Ma, C., Li, G., Han, W., Li, H., Wu, H., Zhao, R., Xie, Y., and Shi, L.}
\newblock Towards artificial general intelligence with hybrid tianjic chip
  architecture.
\newblock {\em Nature 572}, 7767 (2019), pp. 106--111.

\bibitem{rao2021homogeneous}
{\sc Rao, F., and Tao, X.}
\newblock Homogeneous neuromorphic hardware.
\newblock {\em Science 373}, 6561 (2021), pp. 1310--1311.

\bibitem{rao2004hierarchical}
{\sc Rao, R.~P.}
\newblock Hierarchical bayesian inference in networks of spiking neurons.
\newblock {\em Advances in neural information processing systems (Adv. Neu.
  Info. Process. Sys) 17\/} (2004), pp. 1113---1120.

\bibitem{roy2019towards}
{\sc Roy, K., Jaiswal, A., and Panda, P.}
\newblock Towards spike-based machine intelligence with neuromorphic computing.
\newblock {\em Nature 575\/} (2019), pp. 607--617.

\bibitem{salvatori2023brain}
{\sc Salvatori, T., Mali, A., Buckley, C.~L., Lukasiewicz, T., Rao, R.~P.,
  Friston, K., and Ororbia, A.}
\newblock Brain-inspired computational intelligence via predictive coding.
\newblock {\em arXiv:2308.07870\/} (2023).

\bibitem{samadi2017deep}
{\sc Samadi, A., Lillicrap, T.~P., and Tweed, D.~B.}
\newblock Deep learning with dynamic spiking neurons and fixed feedback
  weights.
\newblock {\em Neural computation (Neural Comput.) 29}, 3 (2017), 578--602.

\bibitem{santello2009synaptic}
{\sc Santello, M., and Volterra, A.}
\newblock Synaptic modulation by astrocytes via ca2+-dependent glutamate
  release.
\newblock {\em Neuroscience 158}, 1 (2009), pp. 253--259.

\bibitem{schwartz2020green}
{\sc Schwartz, R., Dodge, J., Smith, N.~A., and Etzioni, O.}
\newblock Green {AI}.
\newblock {\em Communications of the ACM 63}, 12 (2020), pp. 54--63.

\bibitem{singh2022sleep}
{\sc Singh, D., Norman, K.~A., and Schapiro, A.~C.}
\newblock Sleep, brain, and cognition special feature: A model of autonomous
  interactions between hippocampus and neocortex driving sleep-dependent memory
  consolidation.
\newblock {\em Proceedings of the national academy of sciences of the united
  states of america (Proc. Natl. Acad. Sci.) 119\/} (2022), e2201795119.

\bibitem{sprekeler2017functional}
{\sc Sprekeler, H.}
\newblock Functional consequences of inhibitory plasticity: homeostasis, the
  excitation-inhibition balance and beyond.
\newblock {\em Current opinion in neurobiology (Curr. Opin. Neurobiol.) 43\/}
  (2017), pp. 198--203.

\bibitem{thomas2013memristor}
{\sc Thomas, A.}
\newblock Memristor-based neural networks.
\newblock {\em Journal of Physics D: Applied Physics (J. Phys. D: Appl. Phys.)
  46\/} (2013), 093001.

\bibitem{van2008visualizing}
{\sc Van~der Maaten, L., and Hinton, G.}
\newblock Visualizing data using t-sne.
\newblock {\em Journal of machine learning research (JMLR) 9\/} (2008), pp.
  2579--2605.

\bibitem{wang2018learning}
{\sc Wang, W., Pedretti, G., Milo, V., Carboni, R., Calderoni, A., Ramaswamy,
  N., Spinelli, A.~S., and Ielmini, D.}
\newblock Learning of spatiotemporal patterns in a spiking neural network with
  resistive switching synapses.
\newblock {\em Science advances (Sci. Adv.) 4}, 9 (2018), eaat4752.

\bibitem{xie2017efficient}
{\sc Xie, X., Qu, H., Liu, G., and Zhang, M.}
\newblock Efficient training of supervised spiking neural networks via the
  normalized perceptron based learning rule.
\newblock {\em Neurocomputing 241\/} (2017), pp. 152--163.

\bibitem{yin2017algorithm}
{\sc Yin, S., Venkataramanaiah, S.~K., Chen, G.~K., Krishnamurthy, R., Cao, Y.,
  Chakrabarti, C., and Seo, J.-s.}
\newblock Algorithm and hardware design of discrete-time spiking neural
  networks based on back propagation with binary activations.
\newblock {\em 2017 IEEE biomedical circuits and systems conference (BioCAS)\/}
  (2017), 1--5.

\bibitem{zhang2018mixup}
{\sc Zhang, H., Cisse, M., Dauphin, Y.~N., and Lopez-Paz, D.}
\newblock mixup: Beyond empirical risk minimization.
\newblock {\em International conference on learning representations (ICLR)
  2018\/} (2018), 1710.09412.

\bibitem{zhao2023cascaded}
{\sc Zhao, G., Wang, T., Li, Y., Jin, Y., Lang, C., and Ling, H.}
\newblock The cascaded forward algorithm for neural network training.
\newblock {\em arXiv:2303.09728\/} (2023).

\bibitem{zheng2022correcting}
{\sc Zheng, Y., Liu, X.~L., Nishiyama, S., Ranganath, C., and O’Reilly,
  R.~C.}
\newblock Correcting the hebbian mistake: Toward a fully error-driven
  hippocampus.
\newblock {\em PLoS computational biology (PLOS Comput. Biol.) 18}, 10 (2022),
  e1010589.

\bibitem{zou2021breaking}
{\sc Zou, X., Xu, S., Chen, X., Yan, L., and Han, Y.}
\newblock Breaking the von neumann bottleneck: architecture-level
  processing-in-memory technology.
\newblock {\em Science china information sciences (Sci. China Inf. Sci.) 64\/}
  (2021), 160404.

\end{thebibliography}
\bibliographystyle{acm}

\clearpage
\section*{Appendix} 

\subsection*{\underline{Layer Size Experiments and Model Ablations}}

We conducted an experiment probing the performance of a CSDP model (the supervised variant) with respect to its complexity. Specifically, we simulated the model with different first hidden LIF layer sizes (the first hidden layer size was selected as the driving experimental variable). The second hidden layer was set to approximately be one \textcolor{black}{fifth of the} size of the first layer in each case (this was a useful ratio empirically found in preliminary experimentation). The model's generalization/test accuracy (ACC) and (binary) reconstruction cross-entropy (BCE) were recorded and the results of this experiment are reported in table S1. 
We observe that at least $4000$-$5000$ neurons in the first layer are needed to obtain good performance (with respect to classification on MNIST). The units added beyond this range did provide further improvement (in terms of generalization) though a diminishing returns trend was observed as model complexity reached the maximal layer size studied. BCE \textcolor{black}{also reasonably} improves up to about \textcolor{black}{$5000$} units in the first layer and effectively levels out (in the range of \textcolor{black}{$131$-$134$ nats}) beyond this.

Furthermore, we performed a small ablation test of our CSDP model on MNIST, investigating the value/effect that certain components had on overall model performance (in terms of generalization accuracy and binary reconstruction cross entropy). table S2 
examines eight possible ablated configurations related to the CSDP-model: with or without the adaptive threshold, with or without the top-down recurrent synapses, and with or without the class-modulating (skip) synapses. With respect to classification, the results show that all three of these \textcolor{black}{are useful elements in  obtaining the best possible performance -- with the adaptive threshold being one of the most important elements (its removal leads to a drop of nearly $2$\%) -- since removing all three leads to a low test accuracy of about $95$\% (as compared to the $97.58$\% of the full model) and high/poor BCE of $159$ nats (as compared to $134$ nats of the full model). However, while a consistent,  overall performance drop appears to be induced by the removal of these three components, this drop is not a very large nor significant. This indicates that the top-down synapses, the additional class-modulating synapses, and the adaptive threshold do not result in the bulk of the CSDP-SNN's base performance -- it is the CSDP-trained feedforward synapses that are responsible for this. We hypothesize that more complex tasks, such as processing time-varying data or learning part-whole hierarchies from sensory input, would better showcase and more strongly demonstrate the value of the top-down recurrent and skip class/context-modulating synapses (the integration of these elements were motivated to provide upper-layer constraints or a top-down bias on the lower layers of neuronal activity as the message passing is carried out across the SNN layers during its iterative processing of the input). This is an aspect of the SNN model architecture, and its relation to more complex tasks and data streams, that we will explore in future work.
}

\subsection*{\underline{Operators and Model Variables / Constants}}

In tables S3 and S4, 
we collect and briefly define several acronyms, abbreviations, key symbols, and operators used throughout the paper. In table S5, 
we list and describe the key variables for CSDP as well as indicating if they are plastic or not. 
\textcolor{black}{Finally, in} table S6, 
we list and describe CSDP model constants. Note that model constants were lightly tuned using a development/validation (held-out) subset of MNIST/K-MNIST of $10000$ samples (randomly sampled without replacement from the original $60,000$ sample training set of a given database). The development subset(s) was (were) used to manually tune/select hyperparameter values.


\noindent 
\textbf{Hyperparameter Sensitivity.} Models are expected to be robust to most hyperparameter values with respect to the neuronal dynamics, provided that they are chosen in reasonable ranges -- such as the sparsity constraint value for the adaptive threshold (which should not be too high to prevent thresholds from jumping up too quickly) and the base voltage threshold (which should not be too high in terms of decivolts so as to reduce the likelihood of spikes). In terms of the plasticity/learning dynamics, the models are expected to be most sensitive to the goodness threshold $\theta_z$ -- we found that somewhat higher values, e.g., $\theta_z = 10$, were more effective. 

One insight uncovered through preliminary experimentation was that  it was important to select the values for the learning rate $\eta_w$ and the synaptic decay factor $\lambda_d$ in terms of the learning setting ``scale'' at which the synaptic updates were occurring. This meant that for larger batch sizes, a higher learning rate (e.g., $0.002$) and a lower decay yielded more effective/robust performance, whereas for smaller batch sizes (approaching online updates with $1-2$ samples), a lower learning rate with a higher decay was more useful/appropriate. Heuristically, we found that, for MNIST and K-MNIST, a ratio of $\lambda_d/\eta_w = 4$ between the decay rate and learning rate proved to be a useful heuristic (meaning that the modeler would only need to select $\eta_w$ if this heuristic ratio is employed).

\subsection*{\underline{Deriving CSDP from the Goodness Contrastive Functional}}

Here, we show how to derive a simplified form of CSDP from the goodness contrastive functional we presented in the main article for the case of using spikes directly.  Computing goodness with emitted neuronal spikes (omitting $t$ for notational clarity/simplicity) is done in the following manner:
\begin{equation}
    p(y_{type}=1; \mathbf{s}^\ell) = 1/\Big( 1 + \exp \big( -\big[( \Sigma^{J_\ell}_{k=1} (\mathbf{s}^\ell_k)^2) - \theta_z \big] \big) \Big) 
\end{equation}
while the contrastive functional to maximize is:
\begin{equation}
    \mathcal{C}(\mathbf{s}^\ell(t),y_{type}) = \Big( y_{type} \log p(y_{type}=1; \mathbf{s}^\ell) + 
    (1 - y^\ell_{type}) \log p(y_{type}=0; \mathbf{s}^\ell) \Big). \label{eqn:contrastive_loss}
\end{equation}
To obtain the update for a synaptic matrix such as $\mathbf{W}^\ell$ (the bottom-up synapses in both the supervised and \textcolor{black}{unsupervised} variants of CSDP), we take the derivative of Equation \ref{eqn:contrastive_loss} with respect to a single synapse $W^\ell_{ij}$ as follows:
\begin{equation}
    \frac{\partial \mathcal{C}(\mathbf{s}^\ell(t),y_{type})}{\partial W^\ell_{ij}} = \frac{\partial \mathcal{C}(\mathbf{s}^\ell(t),y_{type})}{\partial s^\ell_i(t)} \frac{\partial s^\ell_i(t)}{\partial v^\ell_i(t)} \frac{\partial v^\ell_i(t)}{\partial j^\ell_i(t)} \frac{\partial j^\ell_i(t)}{\partial W^\ell_{ij}} . \label{eqn:csdp_chain_rule}
\end{equation}
Going through each local term in the above chain rule expression, and setting $p(y_{type} = 0; \mathbf{s}^\ell) = 1 - p(y_{type} = 1; \mathbf{s}^\ell)$, we obtain the derivative of the functional with respect to the goodness probability itself:
\begin{equation}
    \frac{\partial \mathcal{C}(\mathbf{s}^\ell(t),y_{type})}{\partial p(y_{type} = 1; \mathbf{s}^\ell)} = \frac{y_{type}}{p(y_{type} = 1; \mathbf{s}^\ell)} - \frac{(1 - y_{type})}{1 - p(y_{type} = 1; \mathbf{s}^\ell)}
\end{equation}
followed by the derivative of the goodness probability with respect to the sum of spikes/activities minus the goodness threshold:
\begin{equation}
    \frac{\partial p(y_{type} = 1; \mathbf{s}^\ell)}{\partial (\sum^{J_\ell}_{k=1} (s^\ell_k)^2) - \theta_z} = \frac{\exp\big( -[(\sum^{J_\ell}_{k=1} (s^\ell_k)^2) - \theta_z] \big)}{(1 + \exp\big( -[(\sum^{J_\ell}_{k=1} (s^\ell_k)^2) - \theta_z] \big)^2}
\end{equation}
and finally we obtain the following key local derivatives for post-synaptic neuron $i$:
\begin{equation}
    \frac{\partial (\sum^{J_\ell}_{k=1} (s^\ell_k)^2) - \theta_z}{\partial \mathbf{s}^\ell_i} = 2 s^\ell_i, \quad \frac{\partial v^\ell_i}{\partial j^\ell_i} = \frac{\Delta t}{\tau_m} R_m, \quad \frac{\partial j^\ell_i}{\partial W^\ell_{ij}} = s^{\ell-1}_j
\end{equation}
which are the derivative of the thresholded activity sum with respect to the activities, the derivative of the voltage/membrane potential with respect to the electrical current, and the derivative of the electrical current with respect \textcolor{black}{to} the synapse $W^\ell_{ij}$ that connects pre-synaptic neuron $j$ to post-synaptic neuron $i$. Finally, we choose to apply the straight-through estimator for the derivative of the spike activity of post-synaptic neuron $i$ with respect to its voltage, i.e., $\partial s^\ell_i/\partial v^\ell_i = 1$ (this does mean a biased estimator of the derivative is being used to make the derivation simpler; we remark that one could alternatively employ a more accurate surrogate function for this derivative as is common in the brain-inspired computing literature).

Given the above local terms, we may rewrite the chain rule equation (Equation \ref{eqn:csdp_chain_rule}) and compose the final update rule, for positive samples, in the following manner:
\begin{equation}
   \frac{\partial \mathcal{C}(\mathbf{s}^\ell,y_{type})}{\partial W^\ell_{ij}}\bigg|_{y_{type}=1} = \overbrace{
   \bigg( \frac{2 R_m \Delta t}{\tau_m} \bigg( \frac{y_{type}}{p(y_{type} = 1; \mathbf{s}^\ell)} \bigg) \bigg( \frac{\exp(-\xi)}{(1 + \exp(-\xi))^2} \bigg)  \bigg)}^{\text{Positive Modulator } \delta^\ell_{i,\text{pos}}}  \overbrace{\bigg( s^\ell_i s^{\ell-1}_j \bigg)}^{\text{Hebbian term}} ,  
\end{equation}
\begin{equation}
   \frac{\partial \mathcal{C}(\mathbf{s}^\ell,y_{type})}{\partial W^\ell_{ij}}\bigg|_{y_{type}=1} = \delta^\ell_{i,\text{pos}} \bigg( s^\ell_i s^{\ell-1}_j \bigg) \label{eqn:csdp_pos_update}
\end{equation}
where the substitution variable is $\xi = (\sum^{J_\ell}_{k=1} (s^\ell_k)^2) - \theta_z $. For negative samples, the final update rule would be:
\begin{equation}
   \frac{\partial \mathcal{C}(\mathbf{s}^\ell,y_{type})}{\partial W^\ell_{ij}}\bigg|_{y_{type}=0} = \overbrace{-\bigg( \frac{2 R_m \Delta t}{\tau_m} \bigg( \frac{(1 - y_{type})}{1 - p(y_{type} = 1; \mathbf{s}^\ell)}\bigg) \bigg( \frac{\exp(-\xi)}{(1 + \exp(-\xi))^2} \bigg)  \Bigg)}^{\text{Negative Modulator } \delta^\ell_{i,\text{neg}}}  \overbrace{\bigg( s^\ell_i s^{\ell-1}_j \bigg)}^{\text{Hebbian term}} 
\end{equation}
\begin{equation}
   \frac{\partial \mathcal{C}(\mathbf{s}^\ell,y_{type})}{\partial W^\ell_{ij}}\bigg|_{y_{type}=0} = \delta^\ell_{i,\text{neg}}  \bigg( s^\ell_i s^{\ell-1}_j \bigg) . \label{eqn:csdp_neg_update}
\end{equation}
To obtain a descent-based (minimizing) optimization direction, one can multiply the above derived rules (as well as the contrastive functional in Equation \ref{eqn:contrastive_loss}) by negative one. 

Note that the CSDP rule presented in the main paper modifies \textcolor{black}{the above partial} derivatives (Equations \ref{eqn:csdp_pos_update} and \ref{eqn:csdp_neg_update}) a bit further by replacing the post-synaptic spike vector $\mathbf{s}^\ell$ with the activity trace vector $\mathbf{z}^\ell$ (the coefficients $2 \Delta t / \tau_m$ are omitted because the scaling is handled by the trace's time constant and the optimization rule's learning rate) and further adding a decay (or an anti-Hebbian) term to the synaptic dynamics (to facilitate long-term synaptic depression). 
In effect, if we examine the above derived rules in Equations \ref{eqn:csdp_pos_update} and \ref{eqn:csdp_neg_update} further, we notice that a CSDP update is effectively a product of two terms -- a modulatory signal (all values of which are folded into $\delta^\ell(t)$)  and a Hebbian term (the post-synaptic spike times the pre-synaptic spike). More importantly, if we slightly adjust the above Equations \ref{eqn:csdp_pos_update} and \ref{eqn:csdp_neg_update}, and place back in the time label $t$, to adhere to the context of the main paper's formulation, we get:
\begin{equation}
   \frac{\partial \mathcal{C}(\mathbf{s}^\ell(t),y_{type})}{\partial W^\ell_{ij}}\bigg|_{y_{type}=1} = \overbrace{\delta^\ell_{i,\text{pos}}(t) }^{\text{Positive Modulator}} \overbrace{\bigg( z^\ell_i(t) s^{\ell-1}_j(t - \Delta t) \bigg)}^{\text{\textcolor{black}{Pre-synaptic-driven} STDP term}} , \label{eqn:trace_csdp_pos_update}
\end{equation}
\begin{equation}
   \frac{\partial \mathcal{C}(\mathbf{s}^\ell(t),y_{type})}{\partial W^\ell_{ij}}\bigg|_{y_{type}=0} = \overbrace{\delta^\ell_{i,\text{neg}}(t) }^{\text{Negative Modulator}} \overbrace{\bigg( z^\ell_i(t) s^{\ell-1}_j(t - \Delta t) \bigg)}^{\text{\textcolor{black}{Pre-synaptic-driven} STDP term}} \label{eqn:trace_csdp_neg_update}
\end{equation}
where we see that CSDP's relationship to spike-timing-dependent plasticity (STDP), specifically the trace-based formulation of STDP. CSDP, at its core, is essentially a switch between two modulated \textcolor{black}{pre-synaptic-driven} STDP rules \textcolor{black}{(our use of the descriptor `pre-synaptic-driven' indicates that the STDP update is triggered by pre-synaptic events or spike pulse occurrences)}. For positive samples, CSDP applies a positively-modulated pre-synaptic-driven STDP adjustment whereas, for negative samples, CSDP entails application of a negatively-modulated pre-synaptic-driven STDP adjustment. 

Finally, note that, in the main paper, the modulation vector signal $\delta^\ell(t)$ encapsulates the positive and negative modulation signals multiplied by the post-synaptic trace values, i.e., each value within it can be rewritten as:  $\delta^\ell_i(t) = (\delta^\ell_{i,\text{pos}} y_{type} + \delta^\ell_{i,\text{neg}} (1 - y_{type})) z^\ell_i(t)$, where the binary label $y_{type}$ turns on only one of the two modulatory signals (given it only takes on binary values). 

\subsection*{\underline{Alternative Plasticity Scheme - The Gated-Voltage Rule}} 

An alternative form to the trace-centered rule we used in this study simply replaces the left argument of the local loss function as follows: $\mathcal{C}(\mathbf{v}^\ell(t) \odot \mathbf{s}^\ell(t), y_{type})$. This is what we refer to as the ``gated voltage rule'' which has the advantage of no longer requiring the tracking of a spike activity trace, i.e., it only requires the use of the spike vector at $t$ as a binary multiplicative gate against the current voltage values. However, our goal was to avoid using voltage directly in the plasticity updates since this would mean that the voltage of neuron $i$ (in layer $\ell$) would be used in the synaptic updates for all other neurons in layer $\ell$ ($\forall k \neq i$), which is biologically implausible. 
Furthermore, we found that, in preliminary experiments\textcolor{black}{, this} rule slightly, albeit consistently, underperformed the trace-based form of the loss and we thus do not investigate it further in this work. 

\subsection*{\underline{A Note on Using Class Values as Inputs}}

In the supervised CSDP model of the main paper, we made the modeling choice to represent class labels as a separate input (that gets encoded as a target spike train). This choice of inputting the class is (slightly) different from $($\textit{1}$)$, 
which chose to directly overlay the labels onto the image pixel space. We note that nothing prevents the modeler from doing the same in our framework and this could be done if desired, further saving on a small number of plastic synapses (i.e., $10 \times J_1$ for the first layer of neuronal cells). We particularly chose to separate out the labels from the sensory inputs as we envisioned in future incarnations/generalizations of the CSDP SNN architecture that other types of target inputs could be provided beyond labels, such as outputs from other neuronal circuits, which would then generally not be directly overlaid over the sensory input.

\subsection*{\underline{A Possible Neuromorphic Construction of CSDP Modulation}}

In the main paper, we highlighted that, neurobiologically, one possible instantiation of the key CSDP modulatory signal $\delta^\ell(t)$ (further treated earlier in this supplement as a gated summation of a positive modulatory value $\delta^\ell_{\text{pos}}$ and a negative one $\delta^\ell_{\text{neg}}$), was in the form of astrocytic support. 
Neuromorphically, implementing this could be done by simply integrating into each layer one single additional ``helper'' neuronal unit -- the astrocyte for layer $\ell$. This helper cell's role would be to only sum up the trace values (or spikes, if trace hardware support is not designed) emitted/corresponding to a particular group of neural hardware units using non-plastic memristor synapses, which would connect in a feedforward fashion each neuron in group/layer $\ell$ to the relevant \textcolor{black}{astrocyte} unit.
This same astrocyte would also receive one additional external input value of either one or zero (which indicates whether a processed sample was real or confabulation and, as seen earlier in this supplement, switches between CSDP's positive or negative modulated STDP adjustment) and finally wire back to these same neurons in $\ell$, also with fixed/non-plastic memristor synapses. This fixed recurrent set of fixed synaptic pathways would be used to deposit CSDP's requisite contrastive modulatory signal to each neuron in the group/layer, i.e., a $\delta^\ell_i(t)$ within $\delta^\ell(t)$ would be deposited/back-transmitted to spiking neuronal cell $i$ in layer $\ell$. In fig. 
S1, we visually depict the high-level view of this envisioned astrocytic computational support sub-circuit.

\subsection*{\underline{Full Algorithm Specification for Simulating a CSDP-Adapted SNN}}

In this section, we provide the pseudocode (Algorithm 1; see next page) required for simulating a recurrent SNN, of arbitrary depth, processing an input pattern over an arbitrary-length stimulus window. Note that Algorithm 1 depicts both supervised context-driven and unsupervised variations of our framework.

\begin{algorithm*}[!ht]
\small{\caption{The contrastive-signal-dependent plasticity credit assignment algorithm. \textcolor{MediumSeaGreen}{Lines in green font} depict portions of the simulation code executed if supervised CSDP is to be utilized.}}
\label{algo:csdp}
\footnotesize
\begin{algorithmic}[1]
   \State {\bfseries Input:} sample $(\mathbf{y},\mathbf{x})$ (with $\mathbf{y}=\emptyset$ if unsupervised), SNN parameters $\Theta$, and 
   \State \hspace{0.85cm} virtual data label $y_{type}$ (with value: $1 =$ ``positive'' and $0 =$ ``negative'')
   \State {\bfseries Hyperparameters:} SGD step size $\eta$, stimulus time $T$, integration constant $\Delta t$, time constants $\tau_m$ and $\tau_{tr}$, 
   \State \hspace{2.8cm} adaptive threshold step $\lambda_v$
   \LineComment $\leftarrow$ denotes the overriding of a variable/object, $\Omega()$ is a bounding function
   \Function{Simulate}{$(\mathbf{y},\mathbf{x}, y_{type}), \Theta$}   
   		\For{$t = 1$ to $T$}
       		\State $\mathbf{s}^0(t) \sim \mathcal{B}(1, p=\mathbf{x})$, \; $\mathbf{s}_y(t) = \mathbf{y}$  \Comment Sample sensory input and context to get an input spike at time $t$
       		\For{$\ell = 1$ to $L$} 
                \LineComment Compute the current and voltage components of layer $\ell$
                \If{$\ell < L$}
                    \State $\mathbf{j}^\ell(t) = R_e \big(\mathbf{W}^\ell \cdot \mathbf{s}^{\ell-1}(t)\big) + R_e \big(\mathbf{V}^\ell \cdot \mathbf{s}^{\ell+1}(t)\big) - R_i \big((\mathbf{M}^\ell \odot (1 - \mathbf{I}^\ell)) \cdot \mathbf{s}^\ell(t)\big)$
                \Else 
                    \State $\mathbf{j}^\ell(t) = R_e\big(\mathbf{W}^\ell \cdot \mathbf{s}^{\ell-1}(t)) - R_i\big((\mathbf{M}^\ell \odot (1 - \mathbf{I}^\ell)) \cdot \mathbf{s}^\ell(t)\big)$
                \EndIf
                \If{$\mathbf{y} \neq \emptyset$}
                    \color{MediumSeaGreen}
                    \State $\mathbf{j}^\ell(t) = \mathbf{j}^\ell(t) + R_e\big(\mathbf{B}^\ell \cdot \mathbf{s}_y(t)\big)$ \Comment Apply label context pressure (if supervised)
                \EndIf
                \color{black}
                \State $\mathbf{v}^\ell(t + \Delta t) = \mathbf{v}^\ell(t) + \frac{\Delta t}{\tau_m} \Big(-\mathbf{v}^\ell(t) + \mathbf{j}^\ell(t) \Big)$
                \LineComment Run the spike model given $\mathbf{j}(t)$ and $\mathbf{v}(t)$ (and depolarize $\mathbf{v}(t)$ if applicable)
                \State $\mathbf{s}^\ell(t) = \mathbf{v}^\ell(t + \Delta t) > v^\ell_{thr}$, \; $\mathbf{v}^\ell(t) = \mathbf{v}^\ell(t+ \Delta t) \odot (1 - \mathbf{s}^\ell(t) )$, 
                \State $v^\ell_{thr} = v^\ell_{thr} + \lambda_v \big( (\sum^{J_\ell}_{j=1}$, \; $\mathbf{s}^\ell(t)_j) - 1 \big)$ 
                \LineComment Update the activation trace for layer $\ell$
                \State $\mathbf{z}^\ell_t = \mathbf{s}^\ell_t - \frac{\Delta t}{\tau_{tr}} \mathbf{z}^\ell_t \odot (1 - \mathbf{z}^\ell_t)$
                \LineComment Calculate local synaptic adjustments and update its synaptic efficacies
                \State Get loss $\mathcal{C}(\mathbf{z}^\ell(t),y_{type})$ \& its partial derivative w.r.t. $\mathbf{z}^\ell(t)$, i.e., $\delta^\ell(t) = \frac{\partial \mathcal{L}_{G}(\mathbf{z}^\ell(t), y_{type})}{\partial \mathbf{z}^\ell(t)} $
                \State $\Delta \mathbf{W}^\ell = \big( R_e \delta^\ell(t) \cdot \big( \mathbf{s}^{\ell-1}(t-1) \big)^T \big) + \lambda_d \big( \mathbf{s}^\ell(t) \cdot (1 - \mathbf{s}^{\ell-1}(t))^T \big)$, \; $\mathbf{W}^\ell \leftarrow \Omega \big( \mathbf{W}^\ell - \eta \Delta \mathbf{W}^\ell \big)$
                \State $\Delta \mathbf{V}^\ell = \big( R_e \delta^\ell(t) \cdot \big( \mathbf{s}^{\ell+1}(t-1) \big)^T \big) + \lambda_d \big( \mathbf{s}^\ell(t) \cdot (1 - \mathbf{s}^{\ell+1}(t))^T \big)$, \; $\mathbf{V}^\ell \leftarrow \Omega \big( \mathbf{V}^\ell - \eta \Delta \mathbf{V}^\ell \big)$ 
                \State $\Delta \mathbf{M}^\ell = \big( R_i \delta^\ell(t) \cdot \big( \mathbf{s}^{\ell}(t-1) \big)^T \big) + \lambda_d \big( \mathbf{s}^\ell(t) \cdot (1 - \mathbf{s}^{\ell}(t))^T \big)$, \; $\mathbf{M}^\ell \leftarrow \Omega \big( \mathbf{M}^\ell - \eta \Delta \mathbf{M}^\ell \big)$  
                \If{$\mathbf{y} \neq \emptyset$}
                    \color{MediumSeaGreen}
                    \State $\Delta \mathbf{B}^\ell = \big( R_e \delta^\ell(t) \cdot \big( \mathbf{s}_y(t-1) \big)^T + \lambda_d \big( \mathbf{s}^\ell(t) \big) \cdot (1 - \mathbf{s}^y(t))^T \big)$, \; $\mathbf{B}^\ell \leftarrow \Omega \big( \mathbf{B}^\ell - \eta \Delta \mathbf{B}^\ell \big)$ 
                \EndIf
                \color{black}
                \LineComment Run $\ell$th local predictor, error neurons and adjust its synaptic efficacies
                \State $\mathbf{v}^\ell_\mu(t + \Delta t) = \mathbf{v}^\ell_\mu(t) + \frac{\Delta t}{\tau_m} \Big( -\mathbf{v}^\ell_\mu(t) + R_e (\mathbf{G}^{\ell+1} \cdot \mathbf{s}^{\ell+1}(t)) \Big)$
                \State $\mathbf{s}^\ell_\mu(t) = \mathbf{v}^\ell_\mu(t + \Delta t) > v^\ell_{thr,\mu}$, 
                \State $\mathbf{e}^\ell(t) = \mathbf{s}^\ell_\mu(t) - \mathbf{s}^\ell(t)$  \Comment Calculate the error/mismatch activities
                \State $\Delta \mathbf{G}^\ell = \big( R_e \mathbf{e}^{\ell-1}(t) \big) \cdot (\mathbf{s}^{\ell}(t))^T$, \; $\mathbf{G}^\ell \leftarrow \Omega \big( \mathbf{G}^\ell - \eta \Delta \mathbf{G}^\ell \big)$
        	\EndFor
            \If{$\mathbf{y} \neq \emptyset$}
                \color{MediumSeaGreen}
                \LineComment Calculate classifier outputs and adjust its synaptic efficacies (if supervised)
                \State  $\mathbf{v}_y(t + \Delta t) = \mathbf{v}_y(t) + \frac{\Delta t}{\tau_m} \big( -\mathbf{v}_y(t) + R_e ( \sum^L_{\ell=1} \mathbf{A}^\ell \cdot \mathbf{s}^\ell(t) ) \big)$
                \State $\mathbf{\mu}_y(t) = \mathbf{v}_y(t + \Delta t) > v^y_{thr}$, \quad  $\mathbf{v}_y(t) = \mathbf{v}_y(t+ \Delta t) \odot (1 - \mathbf{\mu}_y(t) )$
                \State $v^y_{thr} = \mathbf{\mu}_y(t) + \lambda_v \big( (\sum^{C}_{j=1} \mathbf{\mu}_{y,j}(t) - 1 \big)$
                \State $\Delta \mathbf{A}^\ell = R_e\big( \mathbf{\mu}_y(t) - \mathbf{s}_y(t) \big) \cdot (\mathbf{s}^\ell(t))^T$, \; $\mathbf{A}^\ell \leftarrow \Omega \big( \mathbf{A}^\ell - \eta \Delta \mathbf{A}^\ell \big), \; \forall \ell = 1...L$ 
                \color{black}
            \Else 
                \State $\{\mathbf{A}^\ell = \emptyset\}^L_{\ell=1}$ \Comment In unsupervised model, these parameters are set to empty/null values
            \EndIf
    	\EndFor
        \State \textbf{Return} $\Theta = \{\mathbf{W}^\ell,\mathbf{V}^\ell,\mathbf{M}^\ell,\mathbf{B}^{\ell}\}^L_{\ell=1} \bigcup \{\mathbf{G}^\ell\}^L_{\ell=1} \bigcup \{\mathbf{A}^\ell\}^L_{\ell=1}$ \Comment Output newly updated model parameters
    \EndFunction
\end{algorithmic}
\end{algorithm*}


\begin{figure}[!t]
  \begin{center}
    \includegraphics[width=0.71\textwidth]{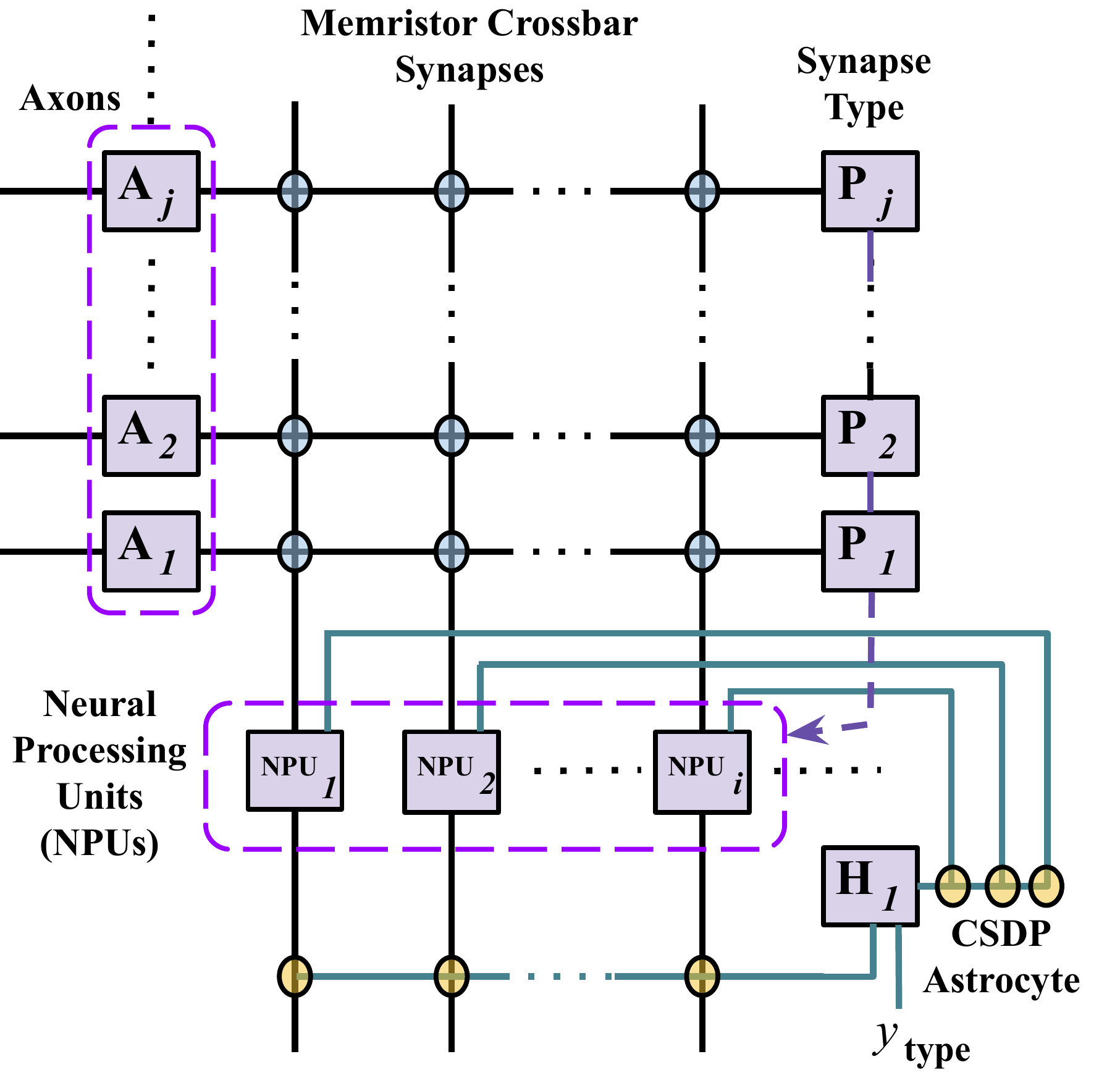}
  \end{center}
  \vspace{-0.2cm}
  \caption{ \textbf{A Neuromorphic Design Sketch of CSDP Astrocytic Computation.} 
  A visualization of a neuromorphic unit implementing the synaptic transmission \textcolor{black}{from} layer $\ell-1$ to layer $\ell$. Hardware axons ($\mathbf{A}_j$) are represented as horizontal lines/rows, hardware dendrites are depicted as vertical lines/columns, and memristive synaptic connections are represented as row-column junctions (light transparent blue circles indicate plastic synaptic junctures while light transparent yellow circles indicate non-plastic synaptic junctures). Neuronal processing units ($\text{NPU}_i$\textcolor{black}{)}, i.e.., hardware leaky integrate-and-fire spiking cells, receive input values from the dendrites while the hardware astrocyte $\mathbf{H}_1$ receives input from a non-plastic excitatory line from the outputs of the NPUs (in this case, their spike values but this could be modified to take in trace signals), as well as \textcolor{black}{a} single binary value indicating whether to operate in positive or negative modulation mode. This astrocyte recurrently emits a value to the NPUs to deposit the required contrastive signal (along an excitatory non-plastic line) to modulate the plastic STDP updates applied to synaptic strengths. Note we also depict, neuromorphically, additional units ($\mathbf{P}_j$) that transmit synaptic ``types'', which \textcolor{black}{are} discrete values representing whether a primary synapse is excitatory (e.g., a one) or inhibitory (e.g., a two) to the NPUs.
  }
  \label{fig:neuromporhic_csdp}
\end{figure}

\begin{table}[!ht]
\begin{center}
\begin{tabular}{l | c | c | c } 
 \hline
  & \multicolumn{2}{|c|}{\textbf{MNIST}} & \\
 \hline
  \textbf{Layer Size} & \textbf{ACC} (\%) & \textbf{BCE} (nats) & \textbf{NoPS} \\
 \hline\hline 
 $N_{\text{lif}} = (500, 100)$ & $93.65 \pm 0.09$ & $162.33  \pm  0.59$ &  $1,206,000$\\
 $N_{\text{lif}} = (1000, 200)$ & $96.13 \pm 0.11$ & $147.15  \pm  1.13$ & $3,232,000$\\
 $N_{\text{lif}} = (2000, 400)$ & $97.21 \pm 0.05$ &  $138.48  \pm  0.41$ & $9,744,000$\\
 $N_{\text{lif}} = (3000, 600)$ & $97.39 \pm 0.05$ & $134.68 \pm 0.49$ & $19,536,000$\\
 $N_{\text{lif}} = (4000, 800)$ & $97.46 \pm 0.07$ & $134.38 \pm 0.12$ & $32,608,000$\\
 $N_{\text{lif}} = (5000, 1000)$ & $97.58 \pm 0.05$ & $134.23 \pm 0.59$ & $48,960,000$ \\
 $N_{\text{lif}} = (6000, 1200)$ & $97.63 \pm 0.04$ & $132.10 \pm 0.31$ & $68,592,000$\\
 $N_{\text{lif}} = (7000, 1400)$ & $97.67 \pm 0.02$ & $131.01 \pm 0.19$ & $91,504,000$\\  
 \hline
\end{tabular}
\caption{ \textbf{CSDP Model Performance with Respect to Number of Neurons.} Measurements, for different size of the hidden layers of the CSDP SNN architecture, of generalization accuracy (ACC, in terms of \%; higher is better) and reconstruction binary cross entropy (BCE, in terms of nats; lower is better) on the MNIST database. Means and standard deviations for $5$ trials are measured for each metric. 
\textcolor{black}{The SNN model training setting was configured to be the same as the one for the model of Table 1 in the main paper ($30$ epochs, batch size of $500$). } 
Note: The number of plastic synapses (NoPS) reported here also includes both classification and reconstruction parameters since reconstruction efficacy were measured. \textcolor{black}{The number of (total) synapses (NoS) would be equal to the NoPS in this case. }
}
\label{table:layer_size_results}
\end{center}
\end{table}

\begin{table}[!ht]
\begin{center}
\begin{tabular}{l | c | c } 
 \hline
  & \multicolumn{2}{c}{\textbf{MNIST}}  \\
 \hline
  \textbf{Model Variant} & \textbf{ACC} (\%) & \textbf{BCE} (nats) \\
 \hline\hline 
 $\lambda_v = $ \xmark, $\mathbf{V}^\ell = $ \xmark, $\mathbf{B}^\ell = $ \xmark & $95.10 \pm 0.42$ & $159.05 \pm 0.31$\\ 
 $\lambda_v =$ \cmark, $\mathbf{V}^\ell = $ \xmark, $\mathbf{B}^\ell = $ \xmark & $96.13 \pm 0.24$ & $140.11 \pm 0.72$\\ 
 $\lambda_v = $ \xmark, $\mathbf{V}^\ell =$ \cmark, $\mathbf{B}^\ell = $ \xmark & $95.97 \pm 0.32$ & $146.02 \pm 0.95$\\
 $\lambda_v = $ \cmark, $\mathbf{V}^\ell =$ \cmark, $\mathbf{B}^\ell = $ \xmark & $95.84 \pm 0.03$ & $147.91 \pm 0.81$\\
 $\lambda_v = $ \xmark, $\mathbf{V}^\ell = $ \xmark,  $\mathbf{B}^\ell = $ \cmark& $95.57 \pm 0.57$ & $141.12 \pm 2.14$\\
 $\lambda_v = $ \cmark, $\mathbf{V}^\ell = $ \xmark, $\mathbf{B}^\ell = $ \cmark& $96.81 \pm 0.05$ & $138.05 \pm 0.39$ \\
 $\lambda_v = $ \xmark, $\mathbf{V}^\ell = $ \cmark,  $\mathbf{B}^\ell = $ \cmark & $95.62 \pm 0.11$ & $153.51 \pm 1.15$\\
 $\lambda_v = $ \cmark, $\mathbf{V}^\ell = $ \cmark, $\mathbf{B}^\ell = $ \cmark & $97.58 \pm 0.05$ & $134.23 \pm 0.59$ \\
 \hline
\end{tabular}
\caption{ \textbf{Ablated CSDP Model Performance.} Measurements, for different ablations of the CSDP SNN architecture, of generalization accuracy (ACC, in terms of \%; higher is better) and reconstruction binary cross entropy (BCE, in terms of nats; lower is better) on the MNIST dataset. 
\textcolor{black}{The architecture and training process of e}ach SNN model examined for this \textcolor{black}{ablation experiment was configured to be the same as the one in Table 1 of the main paper, i.e., $5000$ LIFs in first hidden layer and $1000$ LIFs in the second hidden layer, with training carried out over $30$ epochs with a batch size of $500$ samples.}
}
\label{table:ablation_results}
\end{center}
\end{table}

\begin{table*}[!ht]
\begin{center}
\begin{tabular}{ll} 
 \toprule
 \textbf{Item} & \textbf{Explanation} \\ 
 \midrule
 $\cdot$ & Matrix/vector multiplication \\
 $\odot$ & Hadamard product (element-wise multiplication) \\
 $\mathbf{z}_j$ &  $j$th scalar of vector $\mathbf{z}$ \\
 $||\mathbf{v}||_2$ & Euclidean norm of vector $\mathbf{v}$ \\
 $L$ & Number of layers \\
 $\ell$ &  A specific layer $\ell$ index ($0 \leq \ell \leq L$)\\
 $\mathcal{C}()$ & Local goodness contrastive function \\
 $\delta(t)$ & Goodness/contrastive modulatory signal \\
 $\mathcal{E}()$ & (Global) contrastive functional (goodness over time)\\
 $\mathcal{F}()$ & Optimization objective (contains goodness, reconstruction, classification) \\
 $()^\mathsf{T}$ &  Transpose operation \\
 $\mathbf{W}^\ell$ & A matrix of synaptic weight values \\
 \bottomrule
\end{tabular}
\caption{\textbf{Definitions of mathematical operators and symbols.} Collected definitions of key mathematical symbols and operators used in this work.}
\label{table:symbology}
\end{center}
\end{table*}

\begin{table*}[!ht]
\begin{center}
\begin{tabular}{lll} 
 \toprule
 \textbf{Item} & \textbf{Explanation} \\
 \midrule
 ANN & Artificial neural network \\
 DNN & Deep neural network \\
 Backprop & Backpropagation of errors \\
 NPU & Neural processing unit \\
 BP-FNN & Backprop-trained feedforward network \\
 SNN & Spiking neural network \\
 STDP & Spike-timing-dependent plasticity \\
 CSDP & Contrastive-signal-dependent plasticity\\
 FF & Forward-forward \\
 PFF & Predictive forward-forward\\
 DRTP & Direct random target projection \\
 BFA & Broadcast feedback alignment\\
 L2-SigProp & L2-variant of signal propagation \\
 Loc-Pred & Local predictors\\
 \bottomrule
\end{tabular}
\caption{\textbf{Abbreviation definitions.} Collected definitions of key abbreviations/acronyms used throughout this work.} 
\label{table:terms}
\end{center}
\end{table*}

\begin{table*}[!ht]
\footnotesize
\begin{center}
\begin{tabular}{lll} 
 \toprule
 \textbf{Variable} & \textbf{Explanation} & \textbf{Plastic?}\\ 
 \midrule
 \multicolumn{3}{c}{Leaky Integrator Model Variables} \\
 \hline
 $\mathbf{j}^\ell$ & \begin{tabular}[t]{@{}l@{}}
                        Electrical current input to neurons in layer $\ell$ (vector w/ $J_\ell$ elements)
                    \end{tabular}& N/A (S,U)\\
 $\mathbf{v}^\ell$ & \begin{tabular}[t]{@{}l@{}}
                        Membrane potential of neurons in layer $\ell$ (vector w/ $J_\ell$ elements)
                    \end{tabular}& N/A (S,U)\\
 $\mathbf{s}^\ell$ & \begin{tabular}[t]{@{}l@{}}
                        Spikes emitted from neurons in layer $\ell$ (vector w/ $J_\ell$ elements)
                    \end{tabular}& N/A (S,U)\\
 $\mathbf{z}^\ell$ & \begin{tabular}[t]{@{}l@{}}
                         Traces of neuronal activities in layer $\ell$ (vector w/ $J_\ell$ elements)
                     \end{tabular}& \textcolor{black}{N/A} (S,U)\\
 $\mathbf{v}^\ell_{thr}$ & \begin{tabular}[t]{@{}l@{}}
                         Adaptive voltage thresholds for neurons in layer $\ell$ (vector w/ $J_\ell$ elements)
                     \end{tabular}& Yes (S,U)\\
$\delta(t)$ & \begin{tabular}[t]{@{}l@{}}
                 Modulator (vector) signal for neurons in layer $\ell$ (vector w/ $J_\ell$ elements)
              \end{tabular}& N/A (S,U)\\
\hline
\multicolumn{3}{c}{Synaptic Model Variables} \\
 \hline
 $\mathbf{W}^\ell$ & \begin{tabular}[t]{@{}l@{}}
                        Bottom-up synaptic connections:\\
                        these propagate information from the layer $\ell-1$ below
                    \end{tabular} & Yes (S,U) \\
 $\mathbf{V}^\ell$ & \begin{tabular}[t]{@{}l@{}}
                         Top-down synaptic connections:\\
                         These propagate information from the layer $\ell+1$ above
                     \end{tabular} & Yes (S,U) \\
 $\mathbf{M}^\ell$ & \begin{tabular}[t]{@{}l@{}}
                         Lateral synaptic connections\\
                         These enforce lateral competition among neurons in $\ell$ (cross-layer inhibition)
                     \end{tabular} & Yes (S,U)\\
 $\mathbf{I}^\ell$ & \begin{tabular}[t]{@{}l@{}}
                        Lateral identity masking matrix\\
                        $(1 - \mathbf{I}^\ell)$ is used to enforce main diagonal of $\mathbf{M}^\ell$ to be zero (no self-excitation)
                     \end{tabular} & No (S,U)\\
 $\mathbf{B}^\ell$ & \begin{tabular}[t]{@{}l@{}}
                         Context-mediating synaptic connections:\\
                         These propagate information from (optional) context/class signals
                     \end{tabular} & Yes (S) \\
 $\mathbf{A}^\ell$ & \begin{tabular}[t]{@{}l@{}}
                         Classification synaptic connections:\\
                         These propagate information from a layer $\ell$ to a classification output layer
                     \end{tabular} & Yes (S,U) \\
 $\mathbf{G}^\ell$ & \begin{tabular}[t]{@{}l@{}}
                         Generative synaptic connections:\\
                         These make a local prediction of layer $\ell-1$'s activity given layer $\ell$'s current spikes
                     \end{tabular} & Yes (S,U) \\
 \bottomrule
\end{tabular}
\caption{\textbf{Specification of CSDP model variables.} Model variables, such as neuronal components or synaptic bundles are listed and explained here, with further indication as to whether each is plastic or not. ``S`` denotes if a synaptic bundle variable is applicable only to the supervised variant of CSDP whereas ``U`` indicates if it is applicable to the unsupervised variant.}
\label{table:model_variables}
\end{center}
\end{table*}

\begin{table*}[!ht]
\begin{center}
\begin{tabular}{lll} 
 \toprule
 \textbf{Variable} & \textbf{Explanation} & \textbf{Value}\\ 
 \midrule
 $\Delta t$ & Integration time constant & $3$ ms\\
 $T$ & Simulated time & $90-150$ ms\\
 \hline
 \multicolumn{3}{c}{Leaky Integrator Configuration} \\
 \hline
 $\tau_m$ & Membrane time constant & $100$ ms \\
 $R_E$ & Excitatory membrane resistance & $0.1$ (S), $0.1$ (U) deciOhms \\
 $R_I$ & Inhibitory membrane resistance & $0.035$ (S), $0.01$ (U) deciOhms \\
 $v^\ell_{thr}$ & Voltage threshold & $0.055$ decivolts\\
 $\lambda_v$ & Threshold adaptation factor & $0.001$ decivolts\\
 \hline
 \multicolumn{3}{c}{CSDP Configuration} \\
 \hline
 $\theta_z$ & Goodness threshold & $10$ (S,U) \\
 $\tau_{tr}$ & Variable trace time constant & $13$ ms\\
 $\eta_w$ & Global learning rate (Adam) & $0.002$ \\
 $\lambda_d$ & Synaptic decay & $0.00005$ \\
 $\alpha$ & Negative sample mixing coefficient & $0.5$ (U) \\
 $\mathbf{W}^\ell(0)$ & All initial synapse conditions & $\sim \mathcal{U}(-1,1)$\\ 
 $E$ & Number of epochs/passes through a dataset & $30$ \\
 \bottomrule
\end{tabular}
\caption{\textbf{Specification of CSDP model constants}. Model constants that govern neuronal dynamics, plasticity in CSDP, or general simulation parameters are listed and explained here. ``S`` indicates if a value chosen for a constant applies only to the supervised variant of CSDP while ``U`` denotes if it applies only to the unsupervised variant. Unit of simulation \textcolor{black}{time was} milliseconds (ms).}
\label{table:model_constants}
\end{center}
\end{table*}

\end{document}